\documentclass[10pt]{article} % For LaTeX2e
\usepackage[preprint]{template/tmlr}

\usepackage{amsmath,amsfonts,bm}

\def\eqref#1{equation~\ref{#1}}
\def\1{\bm{1}}

\newcommand{\cov}{\mathrm{cov}}
\newcommand{\RR}{\mathbb{R}}

\newcommand{\EE}{\mathbb{E}}
\newcommand{\BB}{\mathbb{B}}
\newcommand{\vol}{\text{vol}}

\newcommand{\Hess}{\text{Hess}}
\newcommand{\Span}{\text{span}}
\newcommand{\Var}{\text{Var}}
\newcommand{\im}{\text{im}}
\newcommand{\kummer}{{}_1F_1}

\newtheoremstyle{mystyle}%
  {}%
  {}%
  {}%
  {}%
  {\sffamily\bfseries}%
  {.}%
  { }%
  {}%

\renewenvironment{proof}{{\sffamily\bfseries Proof. }}{\qed}

\newtheorem*{remark}{Remark}
\newtheorem{lemma}{Lemma}[subsection]

\declaretheoremstyle[
    headfont=\bfseries, 
    notebraces={[}{]},
    bodyfont=\normalfont,
    mdframed={
        backgroundcolor=Gray!10, 
            linecolor=white!20, 
            innertopmargin=6pt,
            roundcorner=5pt, 
            innerbottommargin=6pt, 
            skipabove=\parsep, 
            skipbelow=\parsep } 
]{definitionstyle}

\declaretheoremstyle[
    headfont=\bfseries, 
    notebraces={[}{]},
    bodyfont=\normalfont,
    mdframed={
        backgroundcolor=Gray!10, 
            linecolor=white!20,
            innertopmargin=6pt,
            roundcorner=5pt, 
            innerbottommargin=6pt, 
            skipabove=\parsep, 
            skipbelow=\parsep } 
]{propositionstyle}

\declaretheoremstyle[
    headfont=\bfseries, 
    notebraces={[}{]},
    bodyfont=\normalfont,
    headpunct={},
    postheadspace=\newline,
    postheadhook={\textcolor{red}{\rule[.6ex]{\linewidth}{0.4pt}}\\},
    spacebelow=\parsep,
    spaceabove=\parsep,
    mdframed={
        backgroundcolor=Gray!20, 
            linecolor=Gray!30, 
            innertopmargin=6pt,
            roundcorner=5pt, 
            innerbottommargin=6pt, 
            skipabove=\parsep, 
            skipbelow=\parsep } 
]{myexamplestyle}

\declaretheorem[
    style=definitionstyle,
    name=Definition,
    numberwithin=section
]{definition}

\usepackage[hidelinks]{hyperref}
\usepackage{url}
\usepackage{subfiles}
\usepackage{placeins}
\usepackage{import}
\usepackage{xifthen}
\usepackage{pdfpages}
\usepackage{transparent}
\usepackage{svg}
\usepackage{calc}
\usepackage{sidecap}
\usepackage[font=small]{caption}
\sidecaptionvpos{figure}{c}

\usepackage[normalem]{ulem}

\title{Identifying latent distances with Finslerian geometry}

\author{\name Alison Pouplin \email alpu@dtu.dk \\
      \addr Technical University of Denmark
      \AND
      \name David Eklund \email david.eklund@ri.se \\
      \addr Research Institutes of Sweden 
      \AND
      \name Carl Henrik Ek \email che29@cam.ac.uk  \\
      \addr University of Cambridge 
      \AND
      \name Søren Hauberg \email sohau@dtu.dk \\
      \addr Technical University of Denmark}

\def\month{10}  % Insert correct month for camera-ready version
\def\year{2023} % Insert correct year for camera-ready version
\def\openreview{\url{https://openreview.net/forum?id=Q2Gi0TUAdS}} % Insert correct link to OpenReview for camera-ready version

\begin{document}

\maketitle

\begin{abstract}

Riemannian geometry provides us with powerful tools to explore the latent space of generative models while preserving the underlying structure of the data. The latent space can be equipped it with a Riemannian metric, pulled back from the data manifold. With this metric, we can systematically navigate the space relying on geodesics defined as the shortest curves between two points.

Generative models are often stochastic, causing the data space, the Riemannian metric, and the geodesics, to be stochastic as well. Stochastic objects are at best impractical, and at worst impossible, to manipulate. A common solution is to approximate the stochastic pullback metric by its expectation. But the geodesics derived from this expected Riemannian metric do not correspond to the expected length-minimising curves.

In this work, we propose another metric whose geodesics explicitly minimise the expected length of the pullback metric. We show this metric defines a Finsler metric, and we compare it with the expected Riemannian metric. In high dimensions, we prove that both metrics converge to each other at a rate of $\mathcal{O}\left(\frac{1}{D}\right)$. This convergence implies that the established expected Riemannian metric is an accurate approximation of the theoretically more grounded Finsler metric. This provides justification for using the expected Riemannian metric for practical implementations.

\end{abstract}

\vspace{0.5cm}
\section{Introduction} 
\label{section:introduction}

\begin{figure}[ht]
    \centering
    \includegraphics[width=\textwidth]{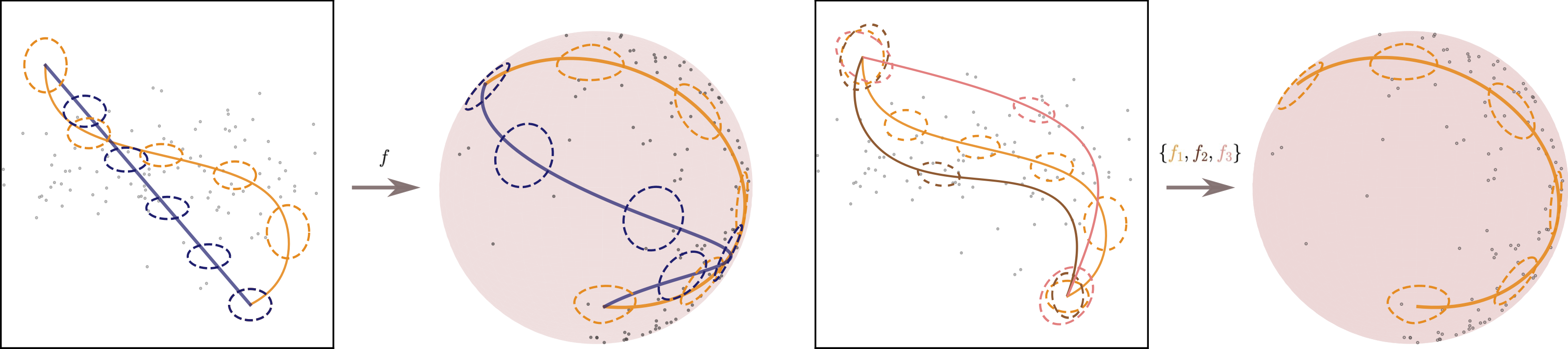}
    \caption{In this illustration, we show the 2-dimensional representation ($\theta, \varphi$) of the sphere parametrised in $\RR^3$ with $f:(\theta, \varphi) \to \cos(\theta)\sin(\varphi), \sin(\theta)\cos(\varphi), \sin(\varphi)$. In practice, we don't have access to well-parametrised manifolds, but instead those manifolds are being shaped by data points in $\RR^D$. The data points are depicted as tiny dots for illustrative purposes. On the left figure, the map $f$ is deterministic, and on the right figure, we imagine that $f$ is stochastic. When we pullback, through $f$, a circle from the sphere to the plane, the circle is deformed to an ellipse. Those ellipses, called indicatrices, are the fingerprints of the metric and they showcase the deformation of the plane. \textbf{Left figure}: In blue, the Euclidean distance, is represented by the straight line in the plane. It is not identifiable and it does not represent a geodesic on the sphere. In orange, the length-minimising curve obtained through the pullback metric of the mapping $f$ leads to a great circle on the sphere. Following the great circle is the fastest way to go from one point to another. Effectively, the pullback metric leads to a geodesic, contrary to the Euclidean metric. Notice that, in the plane, the orange geodesic also follows the direction of the indicatrices, since it is length-minimising.
    \textbf{Right figure}: We imagine that a generative model maps latent space to data space to the data space using a stochastic process $f=\{f_1, f_2, \dots\}$. This leads to a stochastic pullback metric, stochastic indicatrices, and stochastic paths in the latent space.}
    \label{figure:intro_latent}
\end{figure}

Generative models provide a convenient way to learn low-dimensional latent variables $z$ corresponding to data observations $x$ through a smooth function $f: \mathcal{Z}\subset\RR^{q}\to\mathcal{X}\subset\RR^{D}$, such that $x=f(z)$. Through this learnt manifold, one can generate new data or compare observations by interpolating or computing distances. However, doing so by using the Euclidean distance in the latent space is  misleading \citep{hauberg2018only}, because the latent variables are not statistically identifiable. If our observations are lying near a manifold \citep{fefferman:2016}, we want to equip our latent space with a metric that preserves distance measures on it. Figure \ref{figure:intro_latent} (left panel) illustrates the need for defining geometric-aware distances on manifolds. 

Distances on a manifold can be precisely defined using a norm, which is a mathematical function that exhibits several desirable properties such as non-negativity, homogeneity, and the triangle inequality. In particular, a norm can be induced by an inner product (i.e., a quadratic function) that associates each pair of points on the manifold with a scalar value.

To derive the standard Riemannian interpretation of the latent space, we first compute the infinitesimal Euclidean norm according to the data space. Using the Taylor expansion, we have: $\norm{f(z+\Delta z) - f(z)}_{2}^{2} \approx \norm{f(z) + J(z)\Delta z - f(z)}_{2}^{2} = \Delta z^{\top} J(z)^{\top}J(z)\Delta z$. As a first approximation, the norm defined in the latent space locally preserves the Euclidean norm defined in the data space. The curvature of our data manifold is condensed in the Riemannian metric tensor $G_z = J^{\top}(z)J(z)$, which serves as a proxy to define the Riemannian metric: $g_z:(u,v)\to u^{\top}G_z v$. In mathematical jargon, we say that the Riemannian manifold ($\mathcal{Z}$, $g$) is obtained by pulling back the Euclidean metric through the map~$f$. 

% the infinitesimal Euclidean norm in our data space. Using the Taylor expansion, we have: $\norm{f(z+\Delta z) - f(z)}_{2}^{2} \approx \norm{f(z) + J(z)\Delta z - f(z)}_{2}^{2} = \Delta z^{\top} J(z)^{\top}J(z)\Delta z$. As a first approximation, the Euclidean distance can be preserved on $\mathcal{X}$ by defining, locally, a new norm on $\mathcal{Z}$ that includes the term $G = J^{\top}(z)J(z)$, with $z\in\mathcal{Z}$. This term is called a Riemannian metric tensor, and it condenses the geometrical information needed to describe the data space. In mathematical jargon, we say that the manifold $\mathcal{Z}$ is equipped with the Riemannian metric $g:(u,v)\to u^{\top}G v$ obtained by pulling back the Euclidean metric through the immersion $f$. 

Riemannian geometry enables the exploration of the latent space in precise geometric terms, and quantities of interest such as the length, the energy or the volume can be directly derived from the pullback metric. These geometric quantities are, by construction, known to be invariant to reparametrizations of the latent space $\mathcal{Z}$, and are thus statistically identifiable \citep{hauberg2018only}. While this geometric framework exclusively handles deterministic objects, generative models are often stochastic. The learnt map $f$, that mathematically describes those models, is stochastic too. For example, in the celebrated Gaussian Process Latent Variable Model (GPLVM) \citep{lawrence:2003}, this function is a Gaussian process. The pullback metric is then stochastic, and standard differential geometric constructions no longer apply. Navigating the latent manifolds through geodesics (length-minimising curves) becomes practically infeasible. 

% This is shown in the right panel of Figure \ref{figure:intro_latent}. In conclusion, in order to navigate our latent manifold, \textbf{we need a deterministic approximation of our pullback metric.}

% \textbf{In this paper}, we are investigating what could be a good deterministic approximation of the stochastic pullback metric. To navigate a manifold, we are relying on the notion of distances and, more accurately, the notion of norm. In Riemannian geometry, such a norm is induced by an inner product using the Riemannian metric tensor as a proxy quantity. While previous work have used the expectation of the metric tensor as an approximation, we believe it is more sensible to approximate the norm itself. 

Previous research tried to circumvent this problem by approximating the stochastic pullback metric with the expected value of the Riemannian metric tensor, but the derived length is an unintuitive quantities that do not correspond to the expected length:

\[\mathcal{L(\gamma)}\mid_{\EE[G]} := \int \norm{\gamma(t)}_{\EE[G]} \ dt \quad \neq \quad \EE \left[\mathcal{L(\gamma)}\mid_{G} \right] := \int \EE \norm{\gamma(t)}_{G}\ dt.\]

Instead of taking the expectation of the metric tensor, which serves as a surrogate to define a norm, we propose to take the expectation of the norm directly. 

\textbf{In this paper}, we compare our expected norm with the norm induced by the the commonly used expected metric tensor. The main findings are: 

% we believe that the length derived from the pullback metric is a sensible distance measure, and we approximate it by taking its expectation. Then, we \textbf{define a new metric as the expected length derived from the stochastic pullback metric.} This extends the work of \cite{eklund:2019}. 

% The findings are the following: 
\begin{enumerate}[align=left, noitemsep]
    \item The expected norm defines a Finsler metric. Finsler geometry is a generalisation of Riemannian geometry.
    \item For Gaussian processes, the stochastic norm obtained through the pullback metric follows a non-central Nakagami distribution, so our Finsler metric has a closed-form expression.
    \item In high dimensions, for Gaussian processes, our Finsler metric and the previously studied Riemannian metric converge to each other at a rate of $\mathcal{O}\left(\frac{1}{D}\right)$, with $D$ the dimension of the data space. 
\end{enumerate}

\textbf{We conclude} that the Riemannian metric serves as a good approximation of the Finsler metric, which is theoretically more grounded. It further justifies the use of the Riemannian metric in practice when exploring stochastic manifolds. 

\subsection{Outline of the paper}
The paper explores the geometry of latent spaces learned by generative models, which encode a latent low-dimensional manifold that represents observed high-dimensional data. The latent manifold is denoted $\mathcal{Z} \subset \RR^q$ and the data manifold is denoted $\mathcal{X} \subset \RR^D$. 

Assuming the manifold hypothesis holds, we need to define an infinitesimal norm in the latent manifold to compute distances that respect the underlying geometry of the data. Such a norm can be constructed by pulling back the Euclidean distance through the smooth function that maps the latent manifold to the data manifold. This map, $f: \mathcal{Z} \to \mathcal{X}$, mathematically describes the decoder of a trained generative model. Those models being often stochastic, we consider $f$ being a stochastic process. It means that the pullback metric tensor, $G=J^{\top} J$, and its induced norm, $\norm{\cdot}_G: u \to \sqrt{u^{\top} G u}$, are also stochastic. In section \ref{section:expectationOnRandomManifold}, we mathematically define the notion of stochastic pullback metric and stochastic manifolds. 

To circumvent all the challenges posed by this stochastic component, a deterministic approximation of the norm is needed. It can be defined by taking the expectation of the metric tensor. This norm, that we will note $\norm{\cdot}_R: u \to \norm{u}_{\EE[G]}$, has been studied before by \cite{tosi:2014}, and is explained in section \ref{subsection:expectedmetrictensor}. 
In this paper, we propose instead to directly take the expectation of the stochastic norm. This expected norm, noted $\norm{\cdot}_F: u \to \EE\left[\norm{u}_G\right]$, is introduced in section \ref{subsection:expectednorm}. 
The norm $\norm{\cdot}_R$ is defined by a Riemannian metric, and we show that the norm $\norm{\cdot}_F$ defines a Finsler metric. We explain the general difference between Finsler and Riemannian geometry in section \ref{section:comparisonFinslerRiemann}.

The aim of this paper is to compare those two norms. We first draw absolute bounds in section \ref{subsection:absolutebounds}, and then relative bounds in \ref{subsection:relativebounds}. We also investigate the relative difference of the norms when the dimension of the data space increases, in section \ref{subsection:highdimensions}. Finally, we perform some experiments in Section~\ref{section:experiments}, that illustrate the Riemannian and Finsler norm in the same latent space.

\subsection{Related works}
\subsubsection*{Riemannian geometry for machine learning}
When navigating a learnt latent space, the geodesics obtained through the pullback metric not only follow geometrically coherent paths, they also remain invariant under different representations. While two runs of the same model will produce distinct latent manifolds, the geodesics connecting the same chosen points should have the same length. We say that the pullback metric effectively solved the identifiability problem \citep{hauberg2018only}. This has led to the growing adoption of Riemannian geometry in machine learning applications. In robotics, \cite{beik2021learning} used the pullback metric from a variational autoencoder to safely navigate the space of motion patterns, and \cite{scannell2021trajectory} use geodesics under the expected metric in a GPLVM to control quadrotor robot. In proteins modelling, \cite{detlefsen2022learning} showed that the derived geodesics on the space of beta-lactamase follow their phylogenic tree structure. In game content generation, \cite{gonzalez2022mario} generated game levels more coherent and reliable by interpolating data along the geodesics. \cite{jorgensen2021isometric} used the expected Riemannian metric from observations of pairwised distances through a GPLVM to build a probabilistic dimensionality reduction model.

\subsubsection*{Finsler geometry in machine learning}
Our work crucially relies on Finslerian geometry, which has been well-studied mathematically, but has only seen very limited use in machine learning and statistics. We point to two notable exceptions, which are quite distinct from our work.  \cite{lopez:2021} use symmetric spaces to represent graphs and endow these with a Finsler metric to capture dissimilarity structure in the observational data. \cite{ratliff:2021} discuss the role of differential geometry in motion planning for robotics. Along the way, they touch upon Finslerian geometry, but mostly as a neat tool to allow for generalizations. To the best of our knowledge, no prior work has investigated the links between stochastic and Finslerian geometry.

\subsubsection*{Strategies to deal with stochastic Riemannian geometry}
\cite{tosi:2014} and \cite{arvanitidis:iclr:2018} introduced approximation of the pullback metric by taking the expectation of the metric tensor. In those two cases, the map $f$ is respectively a trained Gaussian process, or the decoder of a VAE. In this paper, the derivations only hold if $f$ is a smooth stochastic process (Definition \ref{definition:stochasticprocess}), which is not the case\footnote{the decoder of a VAE, while it decodes to a Gaussian, cannot be considered as a differentiable stochastic process. One reason is because the independence of the probability of the data: $p(x|z) = \prod_{i=1}^{n}p(x_i|z_i)$. Let us assume the opposite: the decoder is a Gaussian process. The covariance of the Gaussian process would be a diagonal matrix because of the independence of the probability of the data. The covariance would correspond to a dirac distribution: $\cov(x_i, x_j) = \delta_{ij}$. However, a stochastic process is differentiable only if the covariance is differentiable, which is not the case of the dirac distribution.} of the VAEs, and hence, our results are not directly applicable to those models.

In addition to the work of \cite{tosi:2014}, a solution to circumvent the randomness of the metric tensor is to consider that the data follows a specific probability distribution. Instead of looking at the shortest path on the data manifold, \cite{arvanatidis:aistats:2022} borrow tools from information geometry and consider the straightest paths on the manifold whose elements are probability distributions.

\section{Expectation on random manifolds} 
\label{section:expectationOnRandomManifold}

The metric pulled back by a stochastic mapping is, de facto, stochastic and endows a random manifold. Unfortunately, we are not yet equipped to derive geometric objects on a random manifold. Instead, we dodge this problem by seeking a deterministic approximation of this stochastic metric. As mentioned above, a common solution is to approximate such a metric by its expectation. In section \ref{subsection:expectedmetrictensor}, we study the expected Riemannian metric. 

The solution suggested by this paper is to approximate the expectation of the lengths instead of the random metric itself. In section \ref{subsection:expectednorm}, we show that this new metric is not Riemannian but Finslerian (Proposition \ref{proposition:definition:finsler}), and it has a closed-form expression when the map $f$ is a Gaussian process (Proposition \ref{proposition:noncentralNakagami}).

\subsection{Random Riemannian geometry}
\label{subsection:randommetric}
The pullback metric is defined as a Riemannian metric if and only if the mapping $f$ is an immersion, which is a differentiable function whose derivatives are injective everywhere on the manifold \citep[Proposition~13.9]{lee:2013}. A manifold equipped with a Riemannian metric is called a Riemannian manifold. \\

\begin{definition}
The pullback of the Euclidean metric through the immersion $f:\mathcal{Z}\to\mathcal{X}$ is a \textbf{Riemannian metric}. It is defined as the inner product $g_z:({\mathcal T_z \mathcal{Z}},{\mathcal T_z \mathcal{Z}})\to \RR_{+} : (u,v)\to u^{\top}G v$, at a specific point $z$ in the manifold $\mathcal{Z}$. $u$ and $v$ are vectors lying in the tangent plane ${\mathcal T_z \mathcal{Z}}$ (ie: the set of all tangent vectors) of the manifold. $G=J^{\top}J$, with $J$ the Jacobian of $f$.
\end{definition}

Since a Riemannian metric is an inner product, it induces a norm: $\norm{\cdot}_G$. We can then define the \textbf{curve length} and \textbf{curve energy} on a manifold: $ L_{G}(\gamma) = \int_0^1 \norm{\dot{\gamma}(t)}_{G} \,dt$ and $E_{G}(\gamma) = \frac{1}{2} \int_0^1 \norm{\dot{\gamma}(t)}_{G}^2 \,dt$, with $\gamma$ a curve defined on $\mathcal{Z}$, and $\dot{\gamma}$ its derivative. A locally length-minimising curve between two connecting points is a \textbf{geodesic}. To obtain a geodesic, we can minimise the curve length, but in practice minimising the curve energy is more efficient. On the manifold, we also define a \textbf{volume measure} in order to integrate probability functions: for $\mathcal{U} \subset \mathcal{Z}$, $\int_{f(\mathcal{U})} h(x) \,dx = \int_{\mathcal{U}} h(f(z)) V_R \,dz$, with $V_R(z) = \sqrt{G_z}$ the volume measure.

% \begin{definition}
% We consider a curve $\gamma$ and its derivative $\dot{\gamma}$ on a Riemannian manifold ($\mathcal{M}$, $g$). Then, we define the \textbf{curve length and curve energy}: 
% \begin{equation*}
% \begin{aligned}
%         L_{G}(\gamma) = \int_0^1 \norm{\dot{\gamma}(t)}_{G} \,dt = \int_0^1 \sqrt{\dot{\gamma}(t)^{\top} G \dot{\gamma}(t)} \,dt,\\ 
%         E_{G}(\gamma) = \int_0^1 \norm{\dot{\gamma}(t)}_{G}^2 \,dt = \int_0^1 \dot{\gamma}(t)^{\top} G \dot{\gamma}(t) \,dt. 
% \end{aligned}
% \end{equation*}
% Locally length-minimising curves between two connecting points are called \textbf{Geodesics}.
% \end{definition}

% Similar to the change of variable theorem, we can define a volume measure to estimate densities on the manifold: \\

% \begin{definition}
% We consider a Riemannian manifold ($\mathcal{M}$, $g$), and $G$ is the corresponding metric tensor. 
% Let $f:\mathcal{M}\to\mathcal{N}$ be an immersion, and $h$ be an integrable function over $\mathcal{N}$. Then, we have:  $\int_{\mathcal{N}} h(y) \,dy = \int_{\mathcal{M}} h \circ f (x) V_R(x)\,dx$, with the  $V_R$ volume measure defined as:
% \[V_R(x) = \sqrt{\det G(x)}\]

% \end{definition}

% All the functionals $L_R$, $E_R$ and $V_R$ can be directly derived from the Riemannian metric tensor $G = J^{\top} J$. 
In addition, we are considering the case where the immersion $f$ is a stochastic process. The outputs of our trained model, $x\in\mathcal{X}$, which represent our data, are random variables. \\

\begin{definition}
\label{definition:stochasticprocess}
    A \textbf{stochastic process} is a collection of random variables $\{X(t, \omega), t \in T \}$ indexed by an index set $T$ defined on a sample space $\Omega$, which represents the set of all possible outcomes. An outcome in $\Omega$ is denoted by $\omega$, and a realisation of the stochastic process is the sequence of $X(\cdot, \omega)$ that depends on the outcome $\omega$. 
\end{definition}

 In this framework, our index set is our latent manifold $T=\mathcal{Z}$, and our sample space $\Omega$ is defined as the set of the model evaluations. For every point $z\in\mathcal{Z}$, every time we execute our model, the output $x=f(z)$ is a random variable following a specific distribution. When the data $x$ follow a Gaussian distribution, the stochastic process is called a \textbf{Gaussian process}. A GP-LVM \citep{lawrence:2003} is a model that learns how to map the data from a latent space to a data space through a Gaussian process. 

When $f$ is a stochastic immersion, the metric tensor becomes a random matrix. In this paper, we call a manifold equipped with the stochastic pullback metric a \textbf{random manifold}, noted $(\mathcal{Z},g)$. As a consequence of the stochastic aspect of the metric, all the functionals are stochastic themselves, and they are no longer trivial to manipulate. \\

\begin{definition}
    A \textbf{random Riemannian metric tensor} is a matrix-valued random field (ie: a collection of matrix-valued random variables $\{G(z, \omega), z \in \mathcal{Z}\}$), whose realisation for a specific evaluation $\omega\in \Omega$ is a Riemannian metric tensor. A \textbf{random Riemannian metric} is a metric induced by a random Riemannian metric tensor: $g_z:({\mathcal T_z \mathcal{Z}},{\mathcal T_z \mathcal{Z}})\to \RR_{+} : (u,v)\to u^{\top}G v$. For the rest of the paper, the associated \textbf{stochastic norm} is noted: 
    \[ 
    \norm{\cdot}_G: {\mathcal T_z \mathcal{Z}} \to \RR_+: u \to \sqrt{g_z(u,u)} \coloneqq \sqrt{u^{\top} G u}
    \]
\end{definition}

If this stochastic norm is induced by $f$ defined as a Gaussian process, then $\norm{\cdot}_G$ follows a non-central Nakagami distribution. This is explained in the proof of Proposition \ref{proposition:noncentralNakagami}.

\subsection{Norm induced by the expected metric tensor}
\label{subsection:expectedmetrictensor}

One way to approximate a random metric tensor is to take its expectation with respect to the collection of random metrics induced by the stochastic process. This has been introduced before by \cite{tosi:2014} GP-LVMs. \\

\begin{definition}
Let $G$ be a stochastic Riemannian metric tensor on the manifold $\mathcal{Z}$. We refer to $\EE[G]$ as the \textbf{expected metric tensor}. It induces a Riemannian metric and a norm on $\mathcal{Z}$. We will note the \textbf{norm induced by the expected metric tensor} as: 
 \[ 
\norm{\cdot}_R: {\mathcal T_z \mathcal{Z}} \to \RR_+: u \to \norm{u}_{\EE[G]} \coloneqq \sqrt{u^{\top} \EE[G] u} \]
\end{definition}

Like any Riemannian metric, we can define the following functionals: $L_R(\gamma) = \int_0^1 \sqrt{\dot{\gamma}(t)^{\top}\EE[G] \dot{\gamma}(t)} \,dt$, $E_R(\gamma) = \int_0^1 \dot{\gamma}(t)^{\top}\EE[G] \dot{\gamma}(t) \,dt = \EE[E_R(\gamma)]$, and $ V_R(z)  = \sqrt{\det \EE[G]}$.

\subsection{Expected paths on random manifolds}
Approximating the stochastic metric by its expectation seems a natural but also ad-hoc solution. If we want to explore a manifold, we might prefer to use a representative quantity, such as the lengths between data points. The expectation of the lengths can give us an idea about how, on average, two points are connected on a random manifold. 
The \textbf{expected curve length}, and its corresponding \textbf{curve energy} on the random manifold $(Z,g)$ are defined as: $L_F(\gamma) = \int_0^1 \EE \left[\sqrt{\dot{\gamma}(t)^{\top} G \dot{\gamma}(t)} \right] \,dt = \EE[L_G(\gamma)]$, and $E_F(\gamma) = \int_0^1 \EE \left[\sqrt{\dot{\gamma}(t)^{\top} G \dot{\gamma}(t) }\right]^2 \,dt$. 

One observation made by \cite{eklund:2019} is that the length ($L_R$) derived from the expected Riemannian metric is not equal to the expected curve length ($L_F$), and their respective energy curves differ by a variance term: 

\begin{equation*}
\begin{aligned}
        E_R(\gamma) - E_F(\gamma) & = \int_0^1 \dot{\gamma}(t)^{\top}\EE[G] \dot{\gamma}(t) - \EE \left[\sqrt{\dot{\gamma}(t)^{\top} G \dot{\gamma}(t) }\right]^2 \,dt \\ 
        & = \int_0^1 \EE\left[\norm{\dot{\gamma}(t)}_G^2\right] - \EE\left[\norm{\dot{\gamma}(t)}_G\right]^2 \,dt = \int_0^1 \Var\left[\norm{\dot{\gamma}(t)}_G\right] \,dt\\
\end{aligned}
\end{equation*}

This term can be regarded as a \textbf{regularisation term} for the Riemannian energy curve: the curve energy $E_R$ might be penalised when the curve goes through regions with high-variance. In practice, for a Gaussian process with a stationary kernel, this variance term is upper bounded by the posterior variance that is relatively low next to the training points and is high outside of the support of the data. Later, we will also see that the functionals agree in high dimensions, leading to the same geodesics (Section \ref{section:comparisonFinslerRiemann}).

\cite{eklund:2019} also noted that these quantities are bounded by the number of dimensions: \\

\begin{restatable}{proposition}{sec31_errorbound} \label{proposition:sec31_errorbound}
\citep{eklund:2019} Let $f:\RR^{q}\to\RR^{D}$ be a stochastic process such that the sequence: $\{f_1', f_2', \dots, f_D'\}$ has uniformly bounded moments. There is then a constant $C$ such that:
\[
0 \leq \frac{L_R - L_F}{L_R}\leq \frac{C}{8D}
\]
\end{restatable}

\subsection{Expected norm and Finsler geometry}
\label{subsection:expectednorm}

Our work builds on \cite{eklund:2019}'s research. We are interested in approximating the stochastic norm instead of the metric tensor, and by doing so, the derived curve length and curve energy are the same ones studied by \cite{eklund:2019}. We go further as we not only compare curve lengths, but the deterministic norms obtained with the stochastic metric. 
\\

\begin{definition}
Let $G$ be a stochastic Riemannian metric tensor on the manifold $\mathcal{Z}$. It induces a stochastic norm, $\norm{\cdot}_G$ on $\mathcal{Z}$. We will note the \textbf{expected norm} as: 
\[ 
\norm{\cdot}_F: {\mathcal T_z \mathcal{Z}} \to \RR_+: u \to \EE[\norm{u}_{G}] \coloneqq \EE\left[\sqrt{u^{\top} G u}\right]
\]
\end{definition}

While it cannot be induced by an inner-product, it is sufficiently convex to be defined as a \textbf{Finsler metric}. \\

\begin{definition}
Let $F:\mathcal{TZ} \rightarrow \mathbb{R}_{+}$ be a continuous non-negative function defined on the tangent bundle $\mathcal{TZ}$ of a differentiable manifold $\mathcal{Z}$. 

We say that $F$ is a \textbf{Finsler metric} if, for each point $z$ of $\mathcal{Z}$ and $v$ on $\mathcal{T}_z \mathcal{Z}$, we have (1) \textbf{Positive homogeneity}: $\forall \lambda \in \mathbb{R}_{+}$, $F(\lambda v) = \lambda F(v)$. (2) \textbf{Smoothness}: $F$ is a $C^{\infty}$ function on the slit tangent bundle $\mathcal{TZ}\setminus{\{0\}}$. (3) \textbf{Strong convexity criterion}: the Hessian matrix $g_{ij}(v) = \frac{1}{2} \frac{\partial^2 F^2}{\partial v^i v^j}(v)$ is positive definite for non-zero $v$.
\end{definition}

A differentiable manifold equipped with a Finsler metric is called a Finsler manifold. Finsler geometry can be seen as an extension of Riemannian geometry, since the requirements for defining a metric are less restrictive. 

\begin{restatable}{proposition}{propDefinitionFinslerOne} \label{proposition:definition:finsler}
 Let $G$ be a stochastic Riemannian metric tensor. Then, the function $F_z: {\mathcal T_z \mathcal{Z}} \rightarrow \RR: u \rightarrow \norm{u}_F $ defines a Finsler metric, but it is not induced by a Riemannian metric.
\end{restatable}

\begin{proof}
If $F$ was induced by a Riemannian metric, then this metric would be defined as: $f_z: \RR^{q} \times \RR^{q} \rightarrow \RR_{+}: (v_1, v_2) \rightarrow \EE[\sqrt{v_1^{\top} G v_2}]^2$. Since a Riemannian metric is an inner product, it should be symmetric, positive, definite and bilinear. Here, we can see that $f_x$ is not bilinear, so $f_z$ is not a Riemannian metric. However, we can prove that $F_z: \RR^{q} \rightarrow \RR: v \rightarrow \EE[\sqrt{v^{\top}  G v}]$ is positive, homogeneous, smooth and strongly convex, and so $F_z$ is a Finsler metric \citep[Definition~2.1]{shen:2016}. For the full proof, see Section \ref{appendix:proof:finslergeometry}.
\end{proof}
\\

So far, we have assumed that $f$ is an immersion and a stochastic process. If we consider $f$ to be a \textbf{Gaussian Process} in particular, the Finsler norm can be rewritten in a closed form expression. \\

\begin{restatable}{proposition}{propDefinitionFinslerTwo} \label{proposition:noncentralNakagami}
Let $f$ be a Gaussian process and $J$ its Jacobian, with $J \sim \mathcal{N}(\EE[J], \Sigma)$.
The Finsler norm can be written as: 
\[
F_z: {\mathcal T_z \mathcal{Z}} \rightarrow \RR_{+}: \norm{v}_F \coloneqq v\to \sqrt{2} \sqrt{v^{\top} \Sigma v}  \frac{\Gamma(\frac{D}{2} + \frac{1}{2})}{\Gamma(\frac{D}{2})}\kummer\left(-\frac{1}{2}, \frac{D}{2}, -\frac{\omega}{2}\right),
\]
with $\kummer$ as the confluent hypergeometric function of the first kind and $\omega= (v^{\top} \Sigma v)^{-1}(v^{\top} \EE[J]^{\top}\EE[J] v)$.
\end{restatable}
\begin{proof}
We suppose that $f$ is a Gaussian process, and so is its Jacobian. $G$ follows a non-central Wishart distribution: $G = J^{\top}J \sim \mathcal{W}_q(D, \Sigma, \Sigma^{-1}\EE[J]^{\top}\EE[J])$. $v^{\top} G v$ is a scalar and also follows a non-central Wishart distribution: $v^{\top}G v \sim \mathcal{W}_1(D, \sigma, \omega)$, with $\sigma = v^{\top} \Sigma v$ and $\omega = (v^{\top} \Sigma v)^{-1}(v^{\top} \EE[J]^{\top}\EE[J] v)$ \citep[Definition~10.3.1]{muirhead:1984}. The square-root of a non-central Wishart distribution follows a non-central Nakagami distribution \citep{hauberg:2018}. Then, by construction, the stochastic norm $\norm{\cdot}_G$ follows a non-central Nakagami distribution. The expectation of this distribution is known, and it has a closed-form expression.
\end{proof}

The confluent hypergeometric function of the first kind, also known as the Kummer function, is a special function that is defined as the solution of a specific second-order linear differential equation. The term $\omega$ appears from the non-central Wishart distribution. When $\omega$ is non-zero, the distribution of the Jacobian shifts away from the origin, and $\omega$ represents the magnitude and the direction of this shift, balanced by the correlation between the variables. In Section \ref{subsection:highdimensions}, to prove our results in high-dimensions, we will assume that our manifold $\mathcal{Z}$ is bounded, and so is $\omega$. 
\section{Comparison of Riemannian and Finsler metrics} 
\label{section:comparisonFinslerRiemann}

\subsection{Theoretical comparison} 

In geometry, we need to define a norm to compute functionals, and the Riemannian metric is conveniently obtained by constructing an inner product. Because of its bilinearity, it greatly simplifies subsequent computations, but it is also restrictive. Relaxing this assumption and  defining a metric as a more general\footnote{The norm actually needs to be strongly convex, but it is not necessary symmetric. This means that, for a vector $v$, we can have a non reversible Finsler metric: $F_x(v) \neq F_x(-v)$. Intuitively, this means that the path used to connect two points would be different depending on the starting point. This asymmetric property becomes valuable when studying the geometry of anisotropic media \citep{markvorsen:2016}, for example. In our case, our Finsler metric is reversible.} norm has been studied by \cite{finsler:1918}, who gave his name to this discipline. 

Finsler geometry is similar to Riemannian geometry without the bilinear assumption, most of the functionals (curve length and curve energy) are defined similarly to those obtained in Riemannian geometry. However, the volume measure is different, and there are at least two definitions of volume measure used in Finsler geometry: the Busemann-Hausdorff volume and the Holmes-Thomson volume measure \citep{wu:2011}. In this paper, we decided to focus on the Busemann-Hausdorff definition (Definition~\ref{definition:busemann-hausdorff}), which is more intuitive and easier to derive. If the Finsler metric is a Riemannian metric, the definition of volume naturally coincides with the Riemannian volume measure.  \\

\begin{definition} \label{definition:busemann-hausdorff}
For a given point $z$ on the manifold, we define the \textbf{Finsler indicatrix} as the set of vectors in the tangent space such that the Finsler metric is equal to: $\{v \in {\mathcal T_z \mathcal{Z}} | F_z(v) = 1\})$. We call $\BB^{n}(1)$ the Euclidean unit ball, and $\vol(\cdot)$ the standard Euclidean volume. In local coordinates ($e^1, \cdots, e^d$) on a Finsler manifold $\mathcal{M}$, the \textbf{Busemann-Hausdorff volume} form is defined as $dV_F = V_F(z) e^1 \wedge \cdots \wedge e^d $, with:
\[V_F(z) = \frac{\vol(\BB^n(1))}{\vol(\{v \in {\mathcal T_z \mathcal{Z}} | F_z(v) < 1\})}.\]
\end{definition}

\begin{figure}[htbp]
    \centering
    \includegraphics[width=\textwidth]{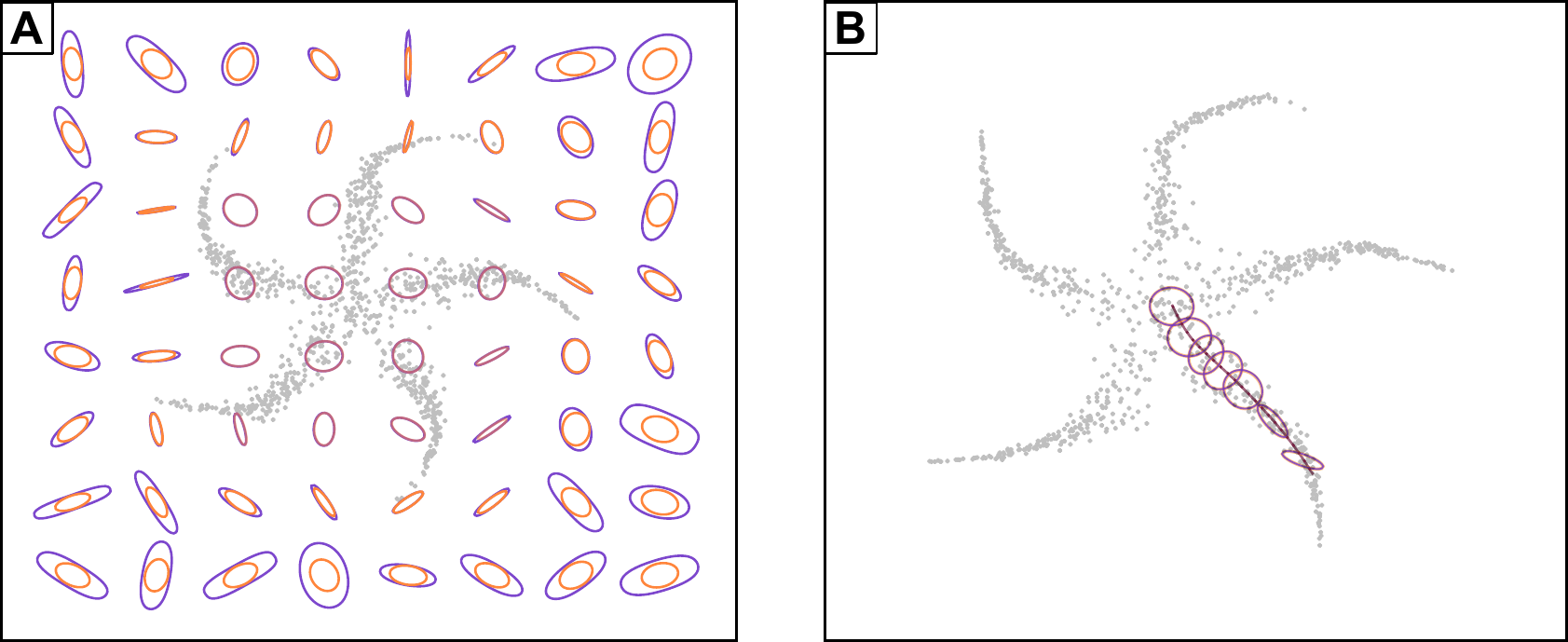}
    \caption{Indicatrice field over the latent space of the pinwheel data (in grey) representing the Riemannian (in orange) and Finslerian (in purple) metrics (See Section~\ref{appendix:experiments}). (A) The indicatrices are computed over a grid in the latent space. (B) The indicatrices are computed along a geodesic: the Riemannian and Finslerian metrics coincide.}
    \label{fig:indicatrices}
\end{figure}

In the definition above, we introduce the notion of \textit{indicatrix}. An indicatrix is a way to represent the distortion induced by the metric on a unit circle. If our metric is Euclidean, we will only have a linear transformation between the latent and the observational spaces, and the indicatrix would still be a circle. Because the Riemannian metric is quadratic, it will always generate an ellipse in the latent space. The Finsler indicatrix, however, would have a convex, even asymmetrical, shape. This difference can be observed in the indicatrix-field represented in Figure~\ref{fig:indicatrices}: The Finsler indicatrices in purple can have almost rectangular shape, while the Riemannian indicatrices, in orange, are ellipses.

There are also a few observations to note in Figure~\ref{fig:indicatrices}. First, in the area of low predictive variance (where data points lie in the latent space), the Finsler and Riemannian indicatrices are alike. This follows from the preceding comment that the metrics diverge by a variance term. If our mapping $f$ was deterministic, both metrics would agree. Second, for every point, the Riemannian indicatrices are always contained by the Finslerian ones, illustrating Proposition~\ref{proposition:absolutebounds} on our absolute bounds in the following section.

\subsection{Absolute bounds on the Finsler metric} \label{subsection:absolutebounds} 

The Finsler norm is upper bounded with the Riemannian norm obtained from the expected metric tensor. It is also lower bounded: \\

\begin{restatable}{proposition}{propabsolutebounds} \label{proposition:absolutebounds}
We define $\alpha = 2 \left(\frac{\Gamma(\frac{D}{2}+\frac{1}{2})}{\Gamma (\frac{D}{2})}\right)^2$. The Finsler norm: $\norm{\cdot}_F$ is bounded by two norms, $\norm{\cdot}_{\alpha\Sigma}$ and $\norm{\cdot}_R$, induced by the two respective Riemannian metric tensors: the covariance tensor $\alpha \Sigma_z$ and the expected metric tensor $\EE[G_z]$.
\[
\forall (z,v) \in \mathcal{Z} \times {\mathcal T_z Z}: \ \norm{v}_{\alpha\Sigma} \leq \norm{v}_F \leq \norm{v}_R 
\]
\end{restatable}
\begin{proof}
The full proof is detailed in Section~\ref{appendix:absolutebounds}, and it can be summarised the following way. The upper bound $ \norm{v}_F \leq \norm{v}_R$, also rewritten as: $  \EE[\sqrt{v^{\top} G v}] \leq \sqrt{v^{\top}\EE[G] v}$, is obtained by applying Jensen's inequality, knowing that the square root $x \rightarrow \sqrt{x}$ is a concave function. The lower bound $\norm{v}_{\alpha\Sigma} \leq \norm{v}_F$, rewritten as $ \sqrt{v^{\top} \alpha  \Sigma v} \leq \EE[\sqrt{v^{\top} G v}]$, is obtained using the closed form expression of the Finsler function.
\end{proof}

The result is illustrated in Figure~\ref{fig:highdims} (lower right). Four metric tensors ($G_1, G_2, G_3, G_4$), each following a non-central Wishart distribution with a specific mean and covariance matrix, have been computed. For each of them, we have drawn the indicatrices ($\{v \in {\mathcal T_z \mathcal{Z}} \ |\  \norm{v} = 1\}$) induced by the norms: $\norm{\cdot}_F$, $\norm{\cdot}_R$ and $\norm{\cdot}_{\alpha\Sigma}$. As expected, we can notice that the $\alpha\Sigma$-indicatrix contains the Finsler indicatrix, itself containing $R$-indicatrix.

By bounding the Finsler metric, we are able to bound their respective functionals: \\
\begin{restatable}{corollary}{coroabsolutebounds} \label{corollary:absolutebounds}
The length, the energy and the Busemann-Hausdorff volume of the Finsler metric are bounded respectively by the Riemannian length, energy and volume of the covariance tensor $\alpha \Sigma$ (noted $L_{\alpha\Sigma}, E_{\alpha\Sigma}, V_{\alpha\Sigma}$) and the expected metric $\EE[G]$ (noted $L_{R}, E_{R}, V_{R}$):
\begin{equation*}
    \begin{aligned}
\forall z \in \mathcal{Z}, \ L_{\alpha\Sigma}(z)  &\leq  L_F(z)  \leq  L_{R}(z) \\
E_{\alpha\Sigma}(z)  &\leq  E_F(z)  \leq E_{R}(z) \\
V_{\alpha\Sigma}(z) &\leq  V_F(z) \leq V_{R}(z) \\
    \end{aligned}
\end{equation*}
\end{restatable}
\begin{proof}
The full proof is detailed in Section~\ref{appendix:absolutebounds}. From Proposition~\ref{proposition:absolutebounds}, we need to integrate each term of the inequality to obtain the length and the energy. The volume is less trivial, since we use the Busemann-Hausdorff definition for measuring $V_F$. We recast the problem in hyperspherical coordinates, and show that the Finsler indicatrix is still bounded.
\end{proof}

\subsection{Relative bounds on the Finsler metric} \label{subsection:relativebounds}

\begin{restatable}{proposition}{propSharpenJensen} \label{proposition:sharpenjensen}
Let $f$ be a stochastic immersion. $f$ induces the stochastic norm $\norm{\cdot}_G$, defined in Section \ref{section:expectationOnRandomManifold}.
The relative difference between the Finsler norm $\norm{\cdot}_F$ and the Riemmanian norm $\norm{\cdot}_R$ is:
\[ 0 \leq \frac{\norm{v}_R -\norm{v}_F}{\norm{v}_R} \leq \frac{\Var\left[\norm{v}_G^2\right]}{2 \EE \left[\norm{v}_G^2\right]^2}. \]
\end{restatable}
\begin{proof}
This proposition is a direct application of the Sharpened Jensen's inequality \citep{liao:2017}.
\end{proof}
% Interestingly, if we were to consider the $f$ as a central Gaussian process, with $J \sim \mathcal{N}(0,\Sigma)$, then the norm would follow a central Nakagami distribution: $\norm{v}_G \sim \text{Nakagami}(m,\Omega)$, with $\Omega = \frac{1}{2} \Sigma^{-1}$ and $m = \frac{D}{2}$. The term $m$ is called a shape parameter, that is also defined as $m = \frac{\EE[\nu]^2}{\Var[\nu]}$, with $\nu \sim \text{Nakagami}(m,\Omega)$. Replacing the upper bound by the shape factor $m$ in the previous equation, we would directly obtain $\frac{1}{D}$ as an upper bound. This is a specific case of Proposition \ref{proposition:relativebounds}, when $\EE[J]=0$.

The previous proposition is valid for any stochastic immersion. We can see that the metrics become equal when the ratio of the variance over the expectation shrinks to zero. This happens in two cases: when the variance converges to zero, which is similar to having a deterministic immersion, and when the number of dimensions increases. The latter case is investigated below for a Gaussian process \footnote{Interestingly, the term $\EE[\nu]^2/\Var[\nu]$, when $\nu$ follows a central Nakagami distribution, is called a shape parameter. It has been introduced by Nakagami himself to study the intensity of fading in radio wave propagation \citep{nakagami1960m}. When $\nu \coloneqq \sqrt{\xi}$ with $\xi \sim \mathcal{W}_1(\Sigma, D)$, then $m = D/2$. This is a particular result obtained from Proposition \ref{proposition:relativebounds}, when $\EE[J]=0$.}.
\\

\begin{restatable}{proposition}{proprelativebounds} \label{proposition:relativebounds}
Let $f$ be a Gaussian process. We note $\omega = (v^{\top} \Sigma v)^{-1}(v^{\top} \EE[J]^{\top}\EE[J]v)$, with $J$ the jacobian of $f$, and $\Sigma$ the covariance matrix of $J$. \\
The relative ratio between the Finsler norm $\norm{\cdot}_F$ and the Riemmanian norm $\norm{\cdot}_R$ is:
\[ 0 \leq \frac{\norm{v}_R -\norm{v}_F}{\norm{v}_R}  \leq \frac{1}{D+\omega} +\frac{\omega}{(D+\omega)^2}. \]
\end{restatable}
\begin{proof}
$v^{\top}G v$ follows a one-dimension non-central Wishart distribution: $v^{\top}G_z v \sim \mathcal{W}_1(D, \sigma, \omega)$, with $\sigma = v^{\top} \Sigma v$ and $\omega = (v^{\top} \Sigma v)^{-1}(v^{\top} \EE[J]^{\top}\EE[J]v)$. We use the theorem of the moments to obtain both the expectation and the variance, which leads us to the result. 
\end{proof}

As we have seen that the metrics are bounded, it is easy to show that the functionals derived from those metrics are also bounded: 

\vspace{0.5cm}
\begin{restatable}{corollary}{cororelativebounds} \label{corollary:relativebounds}
When $f$ is a Gaussian Process, the relative ratio between the length, the energy and the volume of the Finsler norm (noted $L_F,E_F,V_F$) and the Riemannian norm (noted $L_R, E_R, V_R$) is:
\begin{equation*}
    \begin{aligned}
        0 \leq &	\frac{L_R(z) - L_F(z)}{L_R(z)}\leq \max_{v \in {\mathcal T_z \mathcal{Z}}} \left\{\frac{1}{D+\omega} + \frac{\omega}{(D+\omega)^2}\right\} \\
        0 \leq &	\frac{E_R(z) - E_F(z)}{E_R(z)}\leq \max_{v \in {\mathcal T_z \mathcal{Z}}} \left\{\frac{2}{D+\omega} + \frac{1+2\omega}{(D+\omega)^2} + \frac{2\omega}{(D+\omega)^3} + \frac{\omega^2}{(D+\omega)^4} \right\} \\
        0 \leq &	\frac{V_R(z) - V_F(z)}{V_R(z)}\leq 1 - \left(1-\max_{v\in{\mathcal T_z \mathcal{Z}}} \left\{\frac{1}{D+\omega} + \frac{\omega}{(D+\omega)^2} \right\}\right)^{q} \\
    \end{aligned}
\end{equation*}
\end{restatable}

\begin{proof}
We directly use Proposition~\ref{proposition:relativebounds}. To obtain the inequalities with the lengths and the energies, we first multiply all the terms by the Riemannian metric, and we integrate every term. To obtain the inequality with the volume, similarly to Corollary~\ref{corollary:absolutebounds}, we place ourselves in hyperspherical coordinates and bound the radius of the Finsler indicatrix. The full proof is in Section~\ref{appendix:relativebounds}.
\end{proof}

\begin{figure}[htbp]
    \centering
    \includegraphics[width=\textwidth]{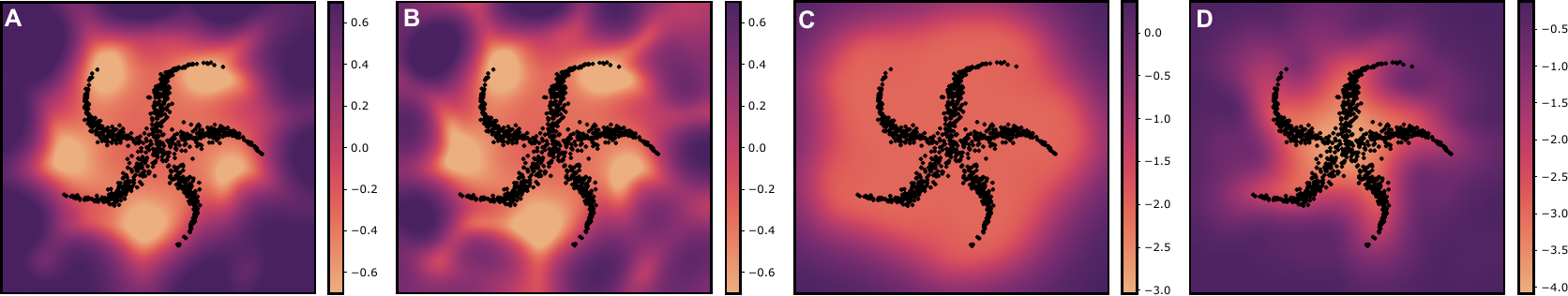}
    \caption{Difference of volume for data embedded in the latent space. (A) Riemannian volume measure, (B) Finslerian (Busemann-Hausdorff) volume measure, (C) Variance of the Gaussian process, (D) Ratio between the Riemannian and Finslerian volume: $(V_R(z) - V_F(z))/ V_R(z)$. All heatmaps are computed in logarithm scale.}
    \label{fig:volume_measures}
\end{figure}

In Figure~\ref{fig:volume_measures}, we can compare the volume measures obtained from the Riemannian and Finsler metrics, and in particular, their ratio in the top right image. When the metrics are computed next to the data points in area where the variance is very low, we can see that the ratio of the volume measure is at the order of magnitude $10^{-4}$. Further away from the data points, the variance increases and so does the difference between the Riemannian and Finsler volume measures. 

\subsection{Results in high dimensions}  \label{subsection:highdimensions}

Proposition \ref{proposition:relativebounds} and Corrollary \ref{corollary:relativebounds} indicate that the metrics become similar when the dimension ($D$) of the observational space increases. If we assume that the latent space is a bounded manifold, the metrics converge to each other at a rate of $\mathcal{O}\left(\frac{1}{D}\right)$, as do their functionals.

We assume that the latent manifold is bounded. Then, we can deduce that (1) the term $\omega$, which represents the non-centrality of the data, does not grow faster than the number of dimensions (See lemma \ref{lemma:omega=O(D)}, in Section \ref{appendix:highdimensions}) and (2) we that the metrics are finite. \\

\begin{restatable}{corollary}{corohighdimfunctionals} \label{corollary:highdimensionsfunctionals}
Let $f$ be a Gaussian Process. In high dimensions, we have: 

\[\frac{L_R(z) - L_F(z)}{L_R(z)}  = \mathcal{O}\left(\frac{1}{D}\right), \quad 
   \frac{E_R(z) - E_F(z)}{E_R(z)} = \mathcal{O}\left(\frac{1}{D}\right), \quad \text{and} \quad
   \frac{V_R(z) - V_F(z)}{V_R(z)} = \mathcal{O}\left(\frac{q}{D}\right). \]
    % \begin{equation*}
    %     \begin{aligned}
    %         \frac{L_R(z) - L_F(z)}{L_R(z)} & = \mathcal{O}\left(\frac{1}{D}\right) \\
    %         \frac{E_R(z) - E_F(z)}{E_R(z)} & = \mathcal{O}\left(\frac{1}{D}\right) \\
    %         \frac{V_R(z) - V_F(z)}{V_R(z)} & = \mathcal{O}\left(\frac{q}{D}\right) \\
    %     \end{aligned}
    % \end{equation*}
    
When $D$ converges toward infinity: $ L_R \underset{+\infty}{\sim} L_F, \   E_R \underset{+\infty}{\sim} E_F$ and $V_R \underset{+\infty}{\sim} V_F$.
\end{restatable}
\begin{proof}
This result follows from Corollary~\ref{corollary:relativebounds}, assuming the latent manifold is bounded. The full proof can be found in Setion~\ref{appendix:highdimensions}.
\end{proof}
\\
\begin{restatable}{corollary}{corohighdimmetrics} \label{corollary:highdimensionsmetrics}
Let $f$ be a Gaussian Process. In high dimensions, the relative ratio between the Finsler norm $\norm{\cdot}_F$ and the Riemmanian norm $\norm{\cdot}_R$ is:
\[ \frac{\norm{v}_R -\norm{v}_F}{\norm{v}_R} = \mathcal{O}\left(\frac{1}{D}\right) \]
    
And, when $D$ converges toward infinity: $ \forall v \in {\mathcal T_z \mathcal{Z}},  \ \norm{v}_R \underset{+\infty}{\sim} \norm{v}_F$.
\end{restatable}
\begin{proof}
Similarly, from Proposition~\ref{proposition:relativebounds}, in a bounded manifold, both metrics converge to each other in high dimensions.
\end{proof}
\begin{figure}[htbp]
    \centering
    \includegraphics[width=\textwidth]{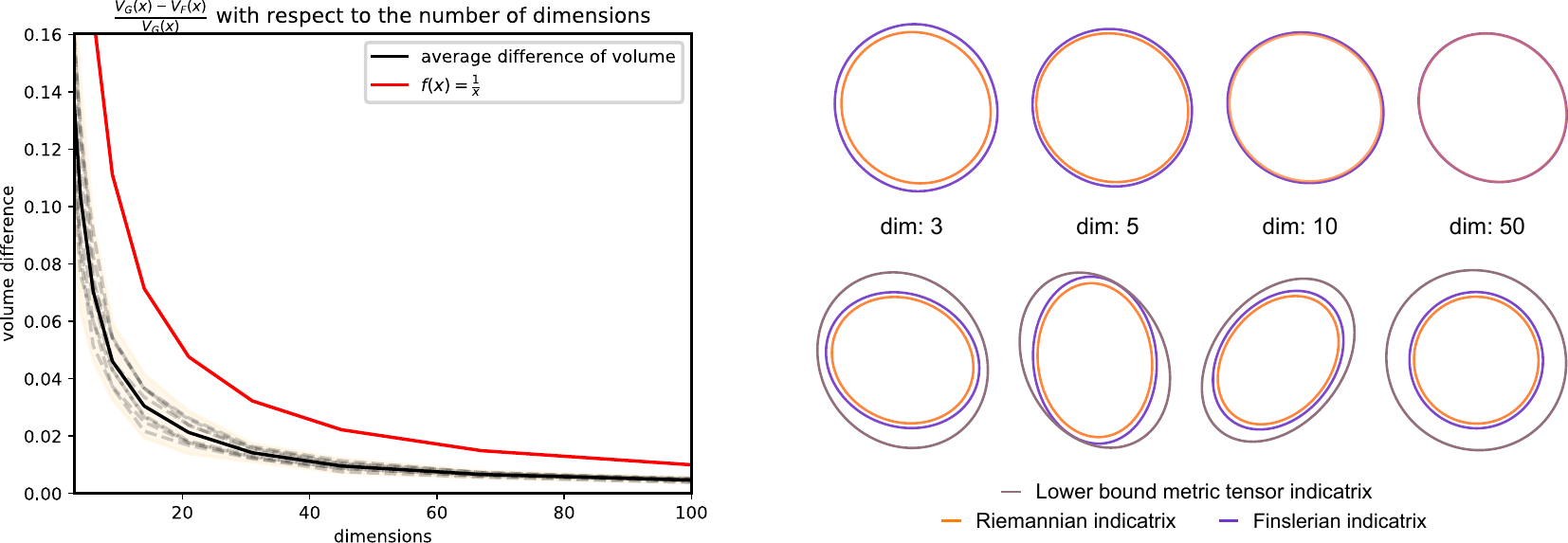}
    \caption{Left: Ratio of volumes $(V_R - V_F)/V_R$ decreasing with respect to the number of dimensions. The results were obtained from using a collection of matrices $\{G_i\}$ following a non-central Wishart distribution. Upper right: The Finsler and Riemannian indicatrices converge towards each other when increasing the number of dimensions. Lower right: Illustration of the absolute bounds in Proposition~\ref{proposition:absolutebounds} with the $\alpha \Sigma$-indicatrices, Riemannian indicatrices and Finsler indicatrices.}
    \label{fig:highdims}
\end{figure}

\section{Experiments} 
\label{section:experiments}

We want to illustrate cases where these metrics differ in practice. For this, we use two synthetic datasets, consisting of pinwheel and concentric circles mapped to a sphere, and four real-world datasets: a font dataset \citep{campbell:2014}, a dataset representing single-cells \citep{Guo:2010}, MNIST \citep{lecun1998mnist} and fashionMNIST \citep{xiao2017fashion}. We trained a GPLVM with and without using stochastic active sets \citep{moreno2022revisiting} to learn a latent manifold. From the learnt model, we can access the Riemannian and Finsler metrics, and minimise their respective curve energies to obtain the corresponding geodesics. All the code has been built using Stochman \citep{software:stochman}, a python library to efficiently compute geodesics on manifolds. 

\subsection{Experiments with synthetic data showing high variance}  \label{subsec:exps:synthetic}
The synthetic data correspond to simple patterns -- a pinwheel and concentric circles -- that have been projected onto a sphere. In those examples, we are plotting curves that does not follow the data points and so, go through regions of high variance. Notably, the background of the latent manifold represents the variance of the posterior distribution in logarithmic scale. For both cases, we have the 3-dimensional data space and the corresponding learnt latent space. The curves are mapped from the latent space to the data space using the forward pass of the GPLVM. 

\begin{figure}[ht]
    \centering
    \includegraphics[width=\textwidth]{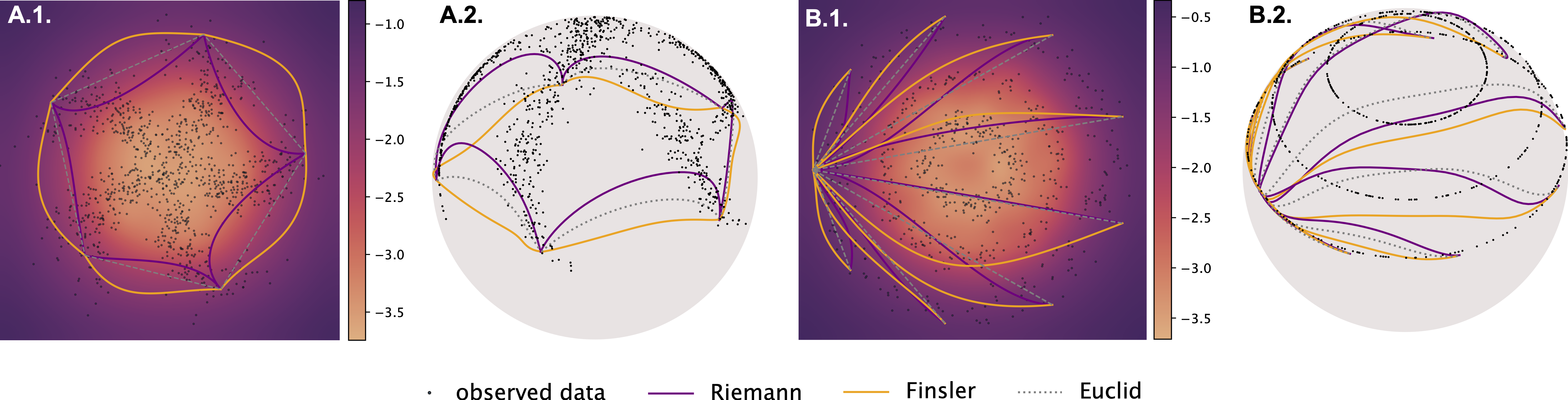}
    \caption{The Riemannian (purple), Finslerian (orange) and Euclidean (dotted gray) geodesics obtained by pulling back the metric through the Gaussian processes of a trained GPLVM. The models, trained on the 3-dimensional synthetic data (Figures A.2, B.2) learnt their latent representations (Figures A.1, B.1)}
    \label{figure:toydata}
\end{figure}

We can see that the Riemannian geodesics tend to avoid area of high variance in both cases: they are attracted to the data, at the detriment of following longer paths in $\RR^3$. The Finslerian geodesics, on the opposite, are not perturbed by the variance term, and will explore regions without any data, following shorter paths in $\RR^3$. The geodesics have been plotted by computing a discretized manifold, obtained from a 10x10 grid with \cite{software:stochman}. This approach proved to be more effective than minimizing the energy along a spline, which was prone to getting trapped in local minima. The GPLVM has been coded in Pyro \citep{bingham:pyro:2019}. All the implementation details can be found in the Appendix~\ref{appendix:experiments}.

\subsection{Experiments with a font dataset and qPCR dataset} \label{subsec:exps:qpcrandfont}
\begin{figure}[ht]
\centering
\includegraphics[width=\textwidth]{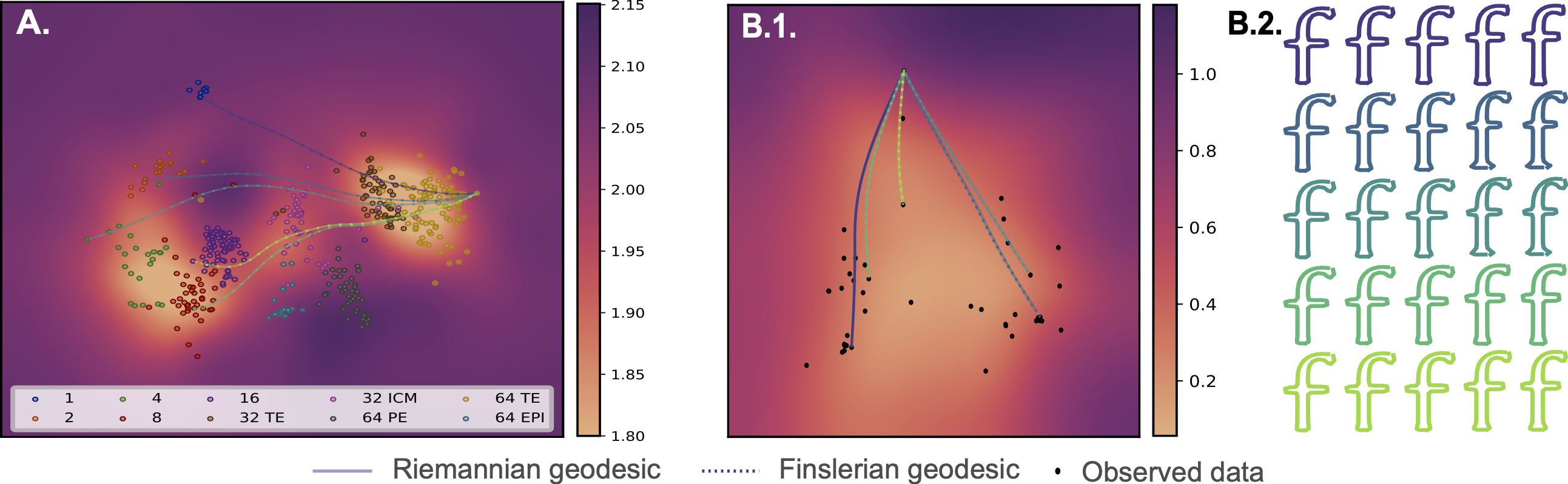}
\caption{The Riemannian (plain line) and Finslerian (dotted line) geodesics obtained by pulling back the metric through the GPLVM. The models trained on single-cells data (which cannot be represented) and font data (Figure B.2) learnt their respective 2-dimensional latent representation (Figures A, B.1).}
\label{fig:exp:qPCR}
\end{figure}

The font dataset \citep{campbell:2014} represents the contour of letters in various font, and the qPCR dataset \citep{Guo:2010} represents single-cells stages. We trained a GPLVM model, also using Pyro \citep{bingham:pyro:2019}, to learn the latent space. From the optimised Gaussian process, we can access the Riemannian and Finsler metric, and minimise their respective curve energies to obtain geodesics. \\
Similar to the previous experiment, the background colour represents the variance of the posterior distribution in logarithmic scale. We can notice that the Riemannian and the Finsler geodesics agrees with each other. This experiment also agrees with our finding that in high dimensions (the dimensions of the single-cells data is 48, the font data is 256), both metrics converge to each other. This concludes that, in practice, the Riemannian metric is a good approximation of the Finsler metric, and Finsler geometry doesn't need to be used in high dimensions.

\subsection{Experiments with MNIST and FashionMNIST} \label{subsec:exps:mnist}

GPLVMs are often hard to train, but they can be scaled effectively using variational inference and inducing points. Using stochastic active sets has been shown to give reliable uncertainty estimates, and the model also learn more meaningful latent representations, as shown in the original paper by \cite{moreno2022revisiting}. For those experiments, we used those models on two well-known benchmarks: MNIST and FashionMNIST.

\begin{figure}[ht]
    \centering
    \includegraphics[width=\textwidth]{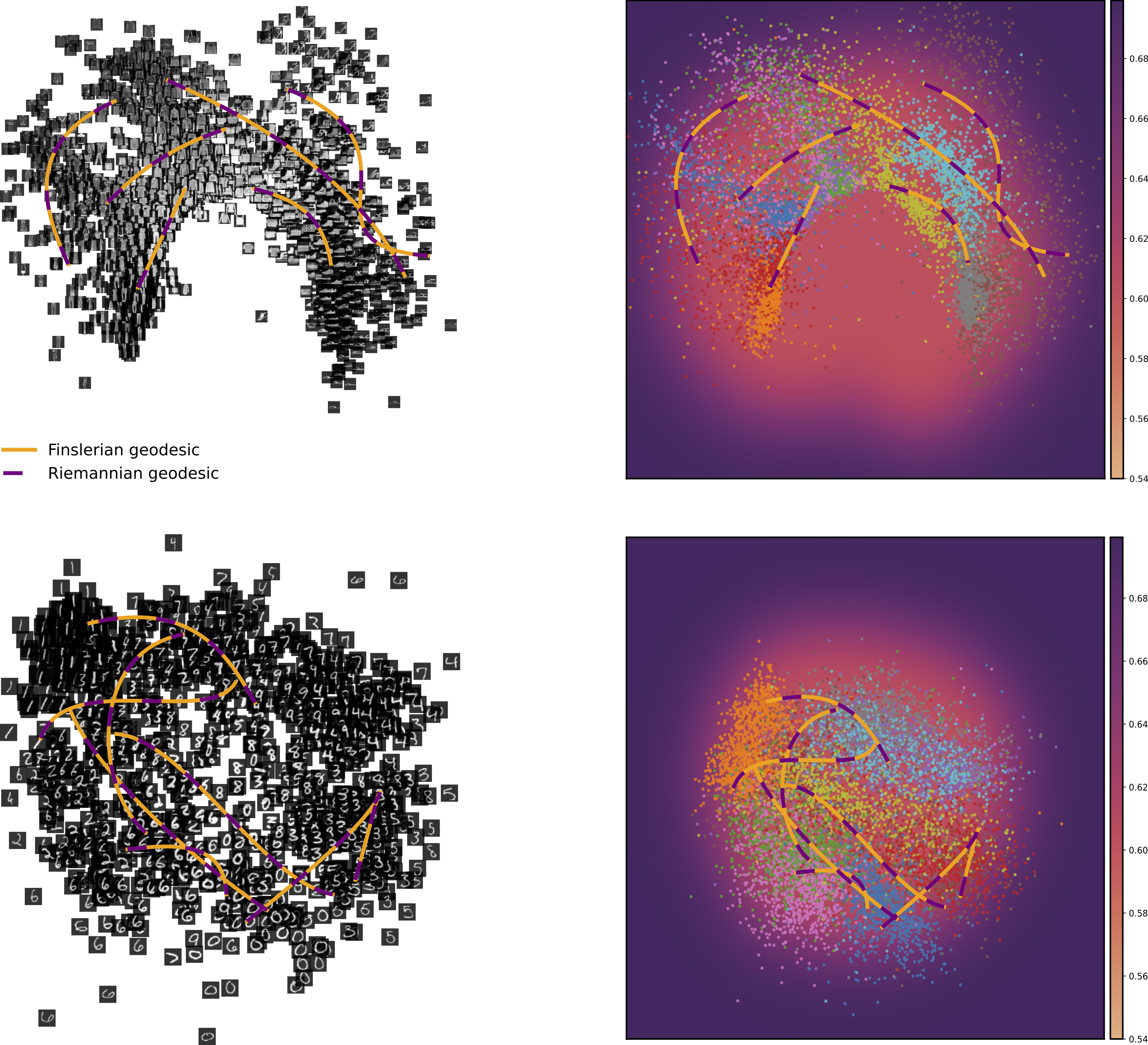}
    \caption{The Riemannian (purple) and the Finslerian (orange) geodesics are plotted. \textbf{Upper figure:} learnt latent space with the images (left) and data points (right) of Fashion MNIST. \textbf{Lower figure:} learnt latent space with the images (left) and data points (right) of  MNIST.}
    \label{figure:fashion_latent}
\end{figure}

In both experiments, because the difference of the variance of the posterior learnt by the GPLVM across the latent manifold is low, as we can notice in the background of the plots on the right, and because the data is high dimensional, we cannot see any difference between the Finslerian and the Riemannian geodesics. Again, in practice, the Riemannian metric is a good approximation of the Finslerian metric.

\section{Discussion} 
\label{section:discussion}

Generative models are often used to reduce data dimension in order to better understand the mechanisms behind the data generating process. We consider the general setting where the mapping from latent variables to observations is driven by a smooth stochastic process, and the sample mappings span Riemannian manifolds. The Riemannian geometry machinery has already been used in the past to explore the latent space.  

In this paper, we have shown how curves and volumes can be identified by defining the length of a latent curve as its expected length measured in the observation space. This is a natural extension of classical differential geometric constructions to the stochastic realm. Surprisingly, we have shown that this does not give rise to a Riemannian metric over the latent space, even if sample mappings do. Rather, the latent representation naturally becomes equipped with a Finsler metric, implying that stochastic manifolds, such as those spanned by Latent Variable Models (LVMs), are inherently more complex than their deterministic counterparts. 

The Finslerian view of the latent representation gives us a suitable general solution to explore a random manifold, but it does not immediately translate into a practical computational tool. As Riemannian manifolds are better understood computationally than Finsler manifolds, we have raised the question: How good an approximation of the Finsler metric can be achieved by a Riemannian metric? The answer turns out to be: quite good. We have shown that as data dimension increases, the Finsler metric becomes increasingly Riemannian. Since LVMs are most commonly applied to high-dimensional data (as this is where dimensionality reduction carries value), we have justification for approximating the Finsler metric with a Riemannian metric such that computational tools become more easily available. In practice we find that geodesics under the Finsler and the Riemannian metric are near identical, except in regions of high uncertainty.

\subsection*{Acknowledgments}
The authors would like to thank Professor Steen Markovsen for the initial discussions on Finsler geometry, and Dr. Cilie Feldager for her invaluable help in taming GP-LVMs. \\
This work was funded in part by the Novo Nordisk Foundation through the Center for Basic Machine Learning Research in Life Science (NNF20OC0062606). It also received funding from the European Research Council (ERC) under the European Union Horizon 2020 research, innovation programme (757360). SH was supported in part by research grants (15334, 42062) from VILLUM FONDEN.

\subsection*{Code availability}
The code used for the experiments in this study is available on GitHub at \url{https://github.com/a-pouplin/latent_distances_finsler}.

\subsection*{Notations}
\begin{tabular}{cp{0.8\textwidth}}
  $\mathcal{Z}, \mathcal{X}$ & Smooth differentiable latent ($\mathcal{Z}$) and data ($\mathcal{X}$) manifold, \\
  $f$ & A stochastic immersion $f: \mathcal{Z}\subset \RR^{q} \to \mathcal{X}\subset \RR^{D}$, \\
  $J$& Jacobian of the stochastic function $f$, \\
  $G$ & Stochastic metric tensor defined as the pullback metric through $f$: $G = J^{\top} J$, \\
  ${\mathcal T_z \mathcal{Z}}$ & Tangent space of the manifold $\mathcal{Z}$ at a point $z$, \\
  $\Sigma$& IF $f$ is a Gaussian process, then $J \sim \prod_{i=1}^{D} \mathcal{N}(\mu_i, \Sigma)$, \\
  $\norm{\cdot}_G$ & Stochastic induced norm: $\norm{v}_G \coloneqq \sqrt{v^{\top} G v}$, \\
  $\norm{\cdot}_R$ & Riemannian induced norm: $\norm{v}_R \coloneqq \sqrt{v^{\top} \EE[G] v} \coloneqq \sqrt{g(v,v)}$, \\
  $\norm{\cdot}_F$ & Finsler norm: $\norm{v}_F \coloneqq \EE[\sqrt{v^{\top} G v}] = F(v)$, \\
  ${L_R}, {E_R}, {V_R}$ & Length, energy and volume obatined from the Riemannian induced norm $\norm{\cdot}_R$, \\
  $L_F, E_F, V_F$ & Length, energy and Busemann Hausdorff volume obtained from the Finsler norm $\norm{\cdot}_F$. \\
\end{tabular}

\bibliography{main}

\begin{thebibliography}{35}
\providecommand{\natexlab}[1]{#1}
\providecommand{\url}[1]{\texttt{#1}}
\expandafter\ifx\csname urlstyle\endcsname\relax
  \providecommand{\doi}[1]{doi: #1}\else
  \providecommand{\doi}{doi: \begingroup \urlstyle{rm}\Url}\fi

\bibitem[Ahmed et~al.(2019)Ahmed, Rattray, and Boukouvalas]{ahmed:2019}
Sumon Ahmed, Magnus Rattray, and Alexis Boukouvalas.
\newblock Grandprix: scaling up the bayesian gplvm for single-cell data.
\newblock \emph{Bioinformatics}, 35\penalty0 (1):\penalty0 47--54, 2019.

\bibitem[Arvanitidis et~al.(2018)Arvanitidis, Hansen, and
  Hauberg]{arvanitidis:iclr:2018}
Georgios Arvanitidis, Lars~Kai Hansen, and S{\o}ren Hauberg.
\newblock {Latent Space Oddity: on the Curvature of Deep Generative Models}.
\newblock In \emph{International Conference on Learning Representations
  (ICLR)}, 2018.

\bibitem[Arvanitidis et~al.(2021)Arvanitidis, González-Duque, Pouplin,
  Kalatzis, and Hauberg]{arvanatidis:aistats:2022}
Georgios Arvanitidis, Miguel González-Duque, Alison Pouplin, Dimitris
  Kalatzis, and Søren Hauberg.
\newblock Pulling back information geometry, 2021.
\newblock URL \url{https://arxiv.org/abs/2106.05367}.

\bibitem[Beik-Mohammadi et~al.(2021)Beik-Mohammadi, Hauberg, Arvanitidis,
  Neumann, and Rozo]{beik2021learning}
Hadi Beik-Mohammadi, S{\o}ren Hauberg, Georgios Arvanitidis, Gerhard Neumann,
  and Leonel Rozo.
\newblock Learning riemannian manifolds for geodesic motion skills.
\newblock \emph{arXiv preprint arXiv:2106.04315}, 2021.

\bibitem[Bingham et~al.(2019)Bingham, Chen, Jankowiak, Obermeyer, Pradhan,
  Karaletsos, Singh, Szerlip, Horsfall, and Goodman]{bingham:pyro:2019}
Eli Bingham, Jonathan~P Chen, Martin Jankowiak, Fritz Obermeyer, Neeraj
  Pradhan, Theofanis Karaletsos, Rohit Singh, Paul Szerlip, Paul Horsfall, and
  Noah~D Goodman.
\newblock Pyro: Deep universal probabilistic programming.
\newblock \emph{The Journal of Machine Learning Research}, 20\penalty0
  (1):\penalty0 973--978, 2019.

\bibitem[Campbell \& Kautz(2014)Campbell and Kautz]{campbell:2014}
Neill D.~F. Campbell and Jan Kautz.
\newblock Learning a manifold of fonts.
\newblock \emph{ACM Trans. Graph.}, 33\penalty0 (4), jul 2014.
\newblock ISSN 0730-0301.
\newblock \doi{10.1145/2601097.2601212}.
\newblock URL \url{https://doi.org/10.1145/2601097.2601212}.

\bibitem[Detlefsen et~al.(2021)Detlefsen, Pouplin, Feldager, Geng, Kalatzis,
  Hauschultz, Duque, Warburg, Miani, and Hauberg]{software:stochman}
Nicki~S. Detlefsen, Alison Pouplin, Cilie~W. Feldager, Cong Geng, Dimitris
  Kalatzis, Helene Hauschultz, Miguel~González Duque, Frederik Warburg, Marco
  Miani, and Søren Hauberg.
\newblock Stochman.
\newblock \emph{GitHub. Note:
  https://github.com/MachineLearningLifeScience/stochman/}, 2021.

\bibitem[Detlefsen et~al.(2022)Detlefsen, Hauberg, and
  Boomsma]{detlefsen2022learning}
Nicki~Skafte Detlefsen, S{\o}ren Hauberg, and Wouter Boomsma.
\newblock Learning meaningful representations of protein sequences.
\newblock \emph{Nature communications}, 13\penalty0 (1):\penalty0 1914, 2022.

\bibitem[Eklund \& Hauberg(2019)Eklund and Hauberg]{eklund:2019}
David Eklund and S{\o}ren Hauberg.
\newblock Expected path length on random manifolds.
\newblock \emph{arXiv preprint arXiv:1908.07377}, 2019.

\bibitem[Fefferman et~al.(2016)Fefferman, Mitter, and
  Narayanan]{fefferman:2016}
Charles Fefferman, Sanjoy Mitter, and Hariharan Narayanan.
\newblock Testing the manifold hypothesis.
\newblock \emph{Journal of the American Mathematical Society}, 29\penalty0
  (4):\penalty0 983--1049, 2016.

\bibitem[Finsler(1918)]{finsler:1918}
Paul Finsler.
\newblock \emph{Ueber kurven und Fl{\"a}chen in allgemeinen R{\"a}umen}.
\newblock Philos. Fak., Georg-August-Univ., 1918.

\bibitem[Gonz{\'a}lez-Duque et~al.(2022)Gonz{\'a}lez-Duque, Palm, Hauberg, and
  Risi]{gonzalez2022mario}
Miguel Gonz{\'a}lez-Duque, Rasmus~Berg Palm, S{\o}ren Hauberg, and Sebastian
  Risi.
\newblock Mario plays on a manifold: Generating functional content in latent
  space through differential geometry.
\newblock In \emph{2022 IEEE Conference on Games (CoG)}, pp.\  385--392. IEEE,
  2022.

\bibitem[Guo et~al.(2010)Guo, Huss, Tong, Wang, {Li Sun}, Clarke, and
  Robson]{Guo:2010}
Guoji Guo, Mikael Huss, Guo~Qing Tong, Chaoyang Wang, Li~{Li Sun}, Neil~D.
  Clarke, and Paul Robson.
\newblock Resolution of cell fate decisions revealed by single-cell gene
  expression analysis from zygote to blastocyst.
\newblock \emph{Developmental Cell}, 18\penalty0 (4):\penalty0 675--685, 2010.
\newblock ISSN 1534-5807.
\newblock \doi{https://doi.org/10.1016/j.devcel.2010.02.012}.
\newblock URL
  \url{https://www.sciencedirect.com/science/article/pii/S1534580710001103}.

\bibitem[Hauberg(2018{\natexlab{a}})]{hauberg2018only}
S{\o}ren Hauberg.
\newblock Only bayes should learn a manifold (on the estimation of differential
  geometric structure from data).
\newblock \emph{arXiv preprint arXiv:1806.04994}, 2018{\natexlab{a}}.

\bibitem[Hauberg(2018{\natexlab{b}})]{hauberg:2018}
S{\o}ren Hauberg.
\newblock {The non-central Nakagami distribution}.
\newblock Technical report, Technical University of Denmark,
  2018{\natexlab{b}}.
\newblock URL
  \url{http://www2.compute.dtu.dk/~sohau/papers/nakagami2018/nakagami.pdf}.

\bibitem[J{\o}rgensen \& Hauberg(2021)J{\o}rgensen and
  Hauberg]{jorgensen2021isometric}
Martin J{\o}rgensen and Soren Hauberg.
\newblock Isometric gaussian process latent variable model for dissimilarity
  data.
\newblock In \emph{International Conference on Machine Learning}, pp.\
  5127--5136. PMLR, 2021.

\bibitem[Kent \& Muirhead(1984)Kent and Muirhead]{muirhead:1984}
John~T. Kent and R.~J. Muirhead.
\newblock {Aspects of Multivariate Statistical Theory.}
\newblock \emph{The Statistician}, 1984.
\newblock ISSN 00390526.
\newblock \doi{10.2307/2987858}.

\bibitem[Kingma \& Ba(2014)Kingma and Ba]{kingma:2014}
Diederik~P Kingma and Jimmy Ba.
\newblock Adam: A method for stochastic optimization.
\newblock \emph{arXiv preprint arXiv:1412.6980}, 2014.

\bibitem[Lawrence(2003)]{lawrence:2003}
Neil Lawrence.
\newblock Gaussian process latent variable models for visualisation of high
  dimensional data.
\newblock \emph{Advances in neural information processing systems}, 16, 2003.

\bibitem[LeCun(1998)]{lecun1998mnist}
Yann LeCun.
\newblock The mnist database of handwritten digits.
\newblock \emph{http://yann. lecun. com/exdb/mnist/}, 1998.

\bibitem[Lee(2013)]{lee:2013}
John~M Lee.
\newblock Smooth manifolds.
\newblock In \emph{Introduction to smooth manifolds}, pp.\  1--31. Springer,
  2013.

\bibitem[Liao \& Berg(2019)Liao and Berg]{liao:2017}
J.~G. Liao and Arthur Berg.
\newblock Sharpening jensen's inequality.
\newblock \emph{The American Statistician}, 73\penalty0 (3):\penalty0 278--281,
  2019.
\newblock \doi{10.1080/00031305.2017.1419145}.

\bibitem[Lopez et~al.(2021)Lopez, Pozzetti, Trettel, Strube, and
  Wienhard]{lopez:2021}
Federico Lopez, Beatrice Pozzetti, Steve Trettel, Michael Strube, and Anna
  Wienhard.
\newblock Symmetric spaces for graph embeddings: A finsler-riemannian approach.
\newblock In Marina Meila and Tong Zhang (eds.), \emph{Proceedings of the 38th
  International Conference on Machine Learning}, volume 139 of
  \emph{Proceedings of Machine Learning Research}, pp.\  7090--7101. PMLR,
  18--24 Jul 2021.
\newblock URL \url{https://proceedings.mlr.press/v139/lopez21a.html}.

\bibitem[Markvorsen(2016)]{markvorsen:2016}
Steen Markvorsen.
\newblock A finsler geodesic spray paradigm for wildfire spread modelling.
\newblock \emph{Nonlinear Analysis: Real World Applications}, 28:\penalty0
  208--228, 2016.

\bibitem[Moreno-Mu{\~n}oz et~al.(2022)Moreno-Mu{\~n}oz, Feldager, and
  Hauberg]{moreno2022revisiting}
Pablo Moreno-Mu{\~n}oz, Cilie Feldager, and S{\o}ren Hauberg.
\newblock Revisiting active sets for gaussian process decoders.
\newblock \emph{Advances in Neural Information Processing Systems},
  35:\penalty0 6603--6614, 2022.

\bibitem[Nakagami(1960)]{nakagami1960m}
Minoru Nakagami.
\newblock The m-distribution—a general formula of intensity distribution of
  rapid fading.
\newblock In \emph{Statistical methods in radio wave propagation}, pp.\  3--36.
  Elsevier, 1960.

\bibitem[Pyro(2022)]{pyro:qPCR}
Pyro.
\newblock Gaussian process latent variable model, 2022.
\newblock URL \url{https://pyro.ai/examples/gplvm.html}.

\bibitem[Rasmussen \& Williams(2005)Rasmussen and
  Williams]{rasmussen:gaussian:2006}
Carl~Edward Rasmussen and Christopher K.~I. Williams.
\newblock \emph{Gaussian Processes for Machine Learning (Adaptive Computation
  and Machine Learning)}.
\newblock The MIT Press, 2005.
\newblock ISBN 026218253X.

\bibitem[Ratliff et~al.(2021)Ratliff, Van~Wyk, Xie, Li, and Rana]{ratliff:2021}
Nathan~D. Ratliff, Karl Van~Wyk, Mandy Xie, Anqi Li, and Muhammad~Asif Rana.
\newblock Generalized nonlinear and finsler geometry for robotics.
\newblock In \emph{2021 IEEE International Conference on Robotics and
  Automation (ICRA)}, pp.\  10206--10212, 2021.
\newblock \doi{10.1109/ICRA48506.2021.9561543}.

\bibitem[Scannell et~al.(2021)Scannell, Ek, and
  Richards]{scannell2021trajectory}
Aidan Scannell, Carl~Henrik Ek, and Arthur Richards.
\newblock Trajectory optimisation in learned multimodal dynamical systems via
  latent-ode collocation.
\newblock In \emph{2021 IEEE International Conference on Robotics and
  Automation (ICRA)}, pp.\  12745--12751. IEEE, 2021.

\bibitem[Shen \& Shen(2016)Shen and Shen]{shen:2016}
Yi-Bing Shen and Zhongmin Shen.
\newblock \emph{Introduction to Modern Finsler Geometry}.
\newblock Co-published with HEP, 2016.
\newblock \doi{10.1142/9726}.
\newblock URL \url{https://www.worldscientific.com/doi/abs/10.1142/9726}.

\bibitem[Tosi et~al.(2014)Tosi, Hauberg, Vellido, and Lawrence]{tosi:2014}
Alessandra Tosi, S{\o}ren Hauberg, Alfredo Vellido, and Neil~D Lawrence.
\newblock Metrics for probabilistic geometries.
\newblock \emph{arXiv preprint arXiv:1411.7432}, 2014.

\bibitem[Wendel(1948)]{wendel:1948}
J.~G. Wendel.
\newblock Note on the gamma function.
\newblock \emph{The American Mathematical Monthly}, 55\penalty0 (9):\penalty0
  563--564, 1948.
\newblock ISSN 00029890, 19300972.
\newblock URL \url{http://www.jstor.org/stable/2304460}.

\bibitem[Wu(2011)]{wu:2011}
Bingye Wu.
\newblock Volume form and its applications in finsler geometry.
\newblock \emph{Publicationes Mathematicae}, 78, 04 2011.
\newblock \doi{10.5486/PMD.2011.4998}.

\bibitem[Xiao et~al.(2017)Xiao, Rasul, and Vollgraf]{xiao2017fashion}
Han Xiao, Kashif Rasul, and Roland Vollgraf.
\newblock Fashion-mnist: a novel image dataset for benchmarking machine
  learning algorithms.
\newblock \emph{arXiv preprint arXiv:1708.07747}, 2017.

\end{thebibliography}
\bibliographystyle{template/tmlr}

\newpage
\appendix
\section{A primer on Geometry} 
\label{section:primerGeometry}
The main purpose of the paper is to define and compare two legitimate metrics to compute the average length between random points. Before going further, it's important to formally define the two metrics (Riemannian and Finsler metrics, respectively) which we do in sections \ref{subsection:riemannianmanifold} and \ref{subsection:finslermanifold}. They are both constructed on topological manifolds, the definition of which is recalled in section \ref{subsection:topologicalmanifold}. We finally introduce the notion of random manifold in section \ref{subsection:randommanifold}, which is the last notion needed to frame our problem of interest: which metric should we use to compute the average distance on a random manifold?

\subsection{Topological and differentiable manifolds} \label{subsection:topologicalmanifold}
This section aims to define core concepts in differential geometry that will be used later to define Riemannian and Finsler manifolds. Recall that two topological spaces are called homeomorphic if there is a continuous bijection between them with continuous inverse.
\begin{definition}
A d-dimensional \textbf{topological manifold} $\mathcal{M}$ is a second-countable Hausdorff topological space such that every point has an open neighbourhood homeomorphic to an open subset of $\RR^d$.
\end{definition}

Let $\mathcal{M}$ be a topological manifold. This means that for any $x \in \mathcal{M}$ there is an open neighbourhood $U_x$ of $x$ and a homeomorphism $\phi_{U_x}: U_x \rightarrow \RR^d$ onto an open subset of $\RR^d$. Suppose that $x,y \in \mathcal{M}$ are such that $U_x \cap U_y \neq \emptyset$, let $U=U_x$, $V=U_y$ and consider the so-called coordinate change map 
\[\phi_{V} \circ \phi^{-1}_{U|\phi_{U}(U \cap V)}: \phi_{U}(U \cap V) \rightarrow \RR^d.\] We call $\mathcal{M}$ together with an open cover $\{U_x\}_{x \in \mathcal{M}}$ as above a \textbf{differentiable} or \textbf{smooth} manifold if the coordinate maps are infinitely differentiable.

Beyond these technical definitions, one can imagine a differentiable manifold as a well-behaved smooth surface that possesses \textit{locally} all the topological properties of a Euclidean space. All the manifolds in this paper are assumed to be differentiable and connected manifolds.

\begin{definition}
We also define, for a differentiable manifold $\mathcal{M}$, the \textbf{tangent space} ${\mathcal T_x M}$ as the set of all the tangent vectors at $x \in \mathcal{M}$, and the \textbf{tangent bundle} $\mathcal{TM}$ the disjoint union of all the tangent spaces: $\mathcal{TM}=\underset{x\in \mathcal{M}}{\cup}{\mathcal T_x M}$. 
\end{definition}

So far, we have only defined topological and differential properties of manifolds. In order to compute geometric quantities, we need to equip those with a metric that helps us derive useful quantities such as lengths, energies and volumes. A metric is a scalar valued function that is defined for each point on the topological manifold and takes as inputs one or two vectors (depending on the type of metric) from the tangent space at the specific point. Such a function can either be defined as a scalar product between two vectors, this is the case of a Riemannian metric or, in the case of a Finsler metric, it is defined similarly to the norm of a vector. We will formally define these metrics and highlight their differences in the following sections.

\subsection{Riemannian manifolds} \label{subsection:riemannianmanifold}

\begin{definition}
Let $\mathcal{M}$ be a manifold. A \textbf{Riemannian metric} is a map assigning at each point $x \in \mathcal{M}$ a scalar product $G(\cdot,\cdot): {\mathcal T_x M} \times {\mathcal T_x M} \rightarrow \RR$, with $G$ a positive definite bilinear map, which is smooth with respect to $x$. A smooth manifold equipped with a Riemannian metric is called a \textbf{Riemannian manifold}. We usually express the metric as a symmetric positive definite matrix $G$, where we have for two vectors $u,v \in {\mathcal T_x M}$: $G(u,v)= \langle u,v\rangle_{G} = u^{\top} G v$. We further define the induced \textbf{norm}: $v \in {\mathcal T_x M}, \norm{v}_G = \sqrt{G(v,v)}$.
\end{definition}

The Riemannian metric here can either refer to the scalar product $G$ itself, or the associated metric tensor $G$. 
% {\color{blue} A good example of a Riemannian metric would be one for the sphere: 1) explain that we can use the parametrisation of the sphere which is an immersion $f$, and 2) the jacobian $J_f$ would be the pushforward of an euclidean space to the sphere.\\}

\begin{figure}[htbp]
    \centering
    \def\svgscale{0.7}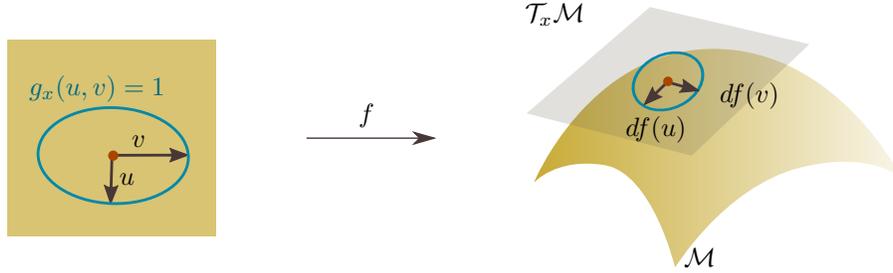
    \caption{$f$ is an immersion that maps a low dimensional manifold to a high dimensional manifold $\mathcal{M}$. On $\mathcal{M}$, a tangent plane $\mathcal{T}_x \mathcal{M}$ is draw at $x$. The indicatrice of the Euclidean metric is plotted in blue. When this metric is pulled-back through $f$, the low dimensional space is now equipped with the pullback metric $g$, which is a Riemannian metric by definition. The vectors $df(u)$ and $df(v)$ are called the push-forwards of the vectors $u$ and $v$ through $f$.}
\end{figure}
\FloatBarrier

\begin{definition}
We consider a curve $\gamma(t)$ and its derivative $\dot{\gamma}(t)$ on a Riemannian manifold $\mathcal{M}$ equipped with the metric $g$. Then, we define the \textbf{length of the curve}: 
\[L_{G}(\gamma) = \int \norm{\dot{\gamma}(t))}_G \,dt = \int \sqrt{g_t(\dot{\gamma}(t),\dot{\gamma}(t))} \,dt,\] where $g_t = g_{\gamma(t)}$. 
Locally length-minimising curves between two connecting points are called \textbf{Geodesics}.
\end{definition}

\begin{definition}
The \textbf{curve energy} is defined as: 
\[E_G(\gamma) = \int \norm{\dot{\gamma}(t))}_G^2 \,dt = \int g_t(\dot{\gamma}(t),\dot{\gamma}(t)) \,dt.\]
\end{definition}

There are two interesting properties to note about the length of a curve and the curve energy. First, the length is parametrisation invariant: for any bijective smooth function $\eta$ on the domain of $\gamma$ we have that $L_{G}(\gamma \circ \eta) = L_{G}(\gamma)$. We also say the Riemannian metric gives us intrinsic coordinates to compute the length. 
Secondly, for a given curve $\gamma$, we have: $L_G(\gamma)^2 \leq 2 E_G(\gamma)$. Because of the invariance of the curve, when we aim to minimise it, a solver can find an infinite number of solutions. On the other hand, the curve energy is convex and will lead to a unique solution. Thus, to obtain a geodesic, instead of solving the corresponding ODE equations, or directly minimising lengths, it is easier in practice to minimise the curve energy, as a minimal energy gives a minimal length.

The Riemannian metric also provides us with an infinitesimal volume element that relates our metric $G$ to an orthonormal basis, the same way the Jacobian determinant accommodate for a change of coordinates in the change of variables theorem.
\begin{definition}
In local coordinates ($e^1, \cdots, e^d$), the \textbf{volume form} of the Riemannian manifold $\mathcal{M}$, equipped with the metric tensor $G$, is defined as: $d V_G = V_G(x) e^1 \wedge \cdots \wedge e^d$, with: 
\[V_G(x) = \sqrt{\det(G)}.\]
\end{definition}
\begin{remark}
The symbol $\wedge$ represents the wedge product and it is used to manipulate differential k-forms. Here, the basis vectors ($e^1, \cdots, e^d$) form a d-dimensional parallelepiped ($e^1 \wedge \cdots \wedge e^d$) with unit volume. 
\end{remark}

\begin{figure}[htbp]
    \centering
    \def\svgscale{0.7}%% Creator: Inkscape 1.0.2 (e86c8708, 2021-01-15), www.inkscape.org
%% PDF/EPS/PS + LaTeX output extension by Johan Engelen, 2010
%% Accompanies image file 'volume_manifold.pdf' (pdf, eps, ps)
%%
%% To include the image in your LaTeX document, write
%%   \input{<filename>.pdf_tex}
%%  instead of
%%   \includegraphics{<filename>.pdf}
%% To scale the image, write
%%   \def\svgwidth{<desired width>}
%%   \input{<filename>.pdf_tex}
%%  instead of
%%   \includegraphics[width=<desired width>]{<filename>.pdf}
%%
%% Images with a different path to the parent latex file can
%% be accessed with the `import' package (which may need to be
%% installed) using
%%   \usepackage{import}
%% in the preamble, and then including the image with
%%   \import{<path to file>}{<filename>.pdf_tex}
%% Alternatively, one can specify
%%   \graphicspath{{<path to file>/}}
%% 
%% For more information, please see info/svg-inkscape on CTAN:
%%   http://tug.ctan.org/tex-archive/info/svg-inkscape
%%
\begingroup%
  \makeatletter%
  \providecommand\color[2][]{%
    \errmessage{(Inkscape) Color is used for the text in Inkscape, but the package 'color.sty' is not loaded}%
    \renewcommand\color[2][]{}%
  }%
  \providecommand\transparent[1]{%
    \errmessage{(Inkscape) Transparency is used (non-zero) for the text in Inkscape, but the package 'transparent.sty' is not loaded}%
    \renewcommand\transparent[1]{}%
  }%
  \providecommand\rotatebox[2]{#2}%
  \newcommand*\fsize{\dimexpr\f@size pt\relax}%
  \newcommand*\lineheight[1]{\fontsize{\fsize}{#1\fsize}\selectfont}%
  \ifx\svgwidth\undefined%
    \setlength{\unitlength}{482.21653242bp}%
    \ifx\svgscale\undefined%
      \relax%
    \else%
      \setlength{\unitlength}{\unitlength * \real{\svgscale}}%
    \fi%
  \else%
    \setlength{\unitlength}{\svgwidth}%
  \fi%
  \global\let\svgwidth\undefined%
  \global\let\svgscale\undefined%
  \makeatother%
  \begin{picture}(1,0.25496594)%
    \lineheight{1}%
    \setlength\tabcolsep{0pt}%
    \put(0,0){\includegraphics[width=\unitlength,page=1]{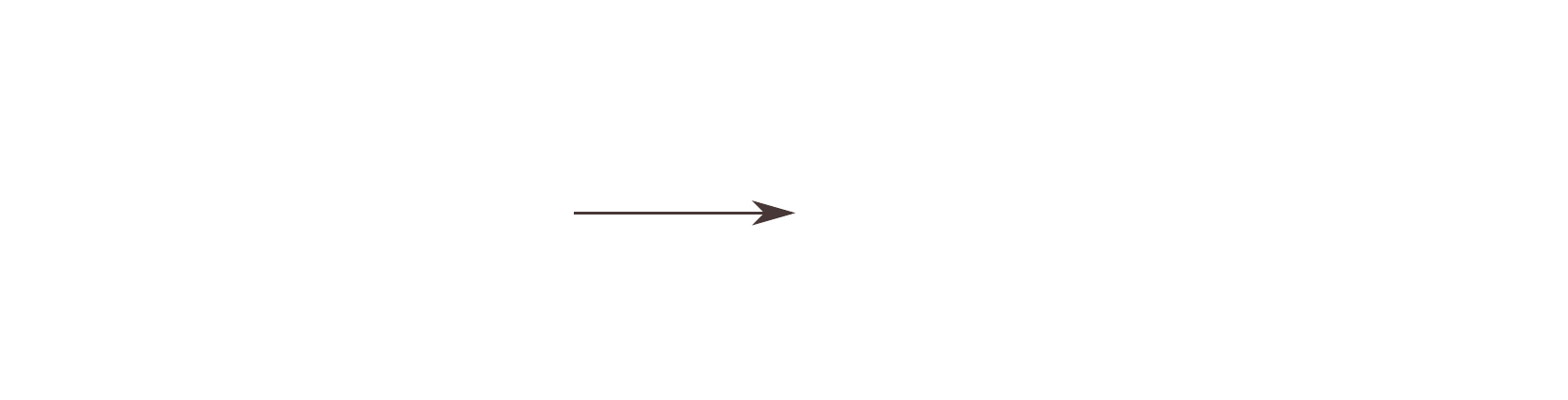}}%
    \put(0.42187623,0.13955699){\makebox(0,0)[lt]{\lineheight{1.25}\smash{\begin{tabular}[t]{l}$f$\end{tabular}}}}%
    \put(0,0){\includegraphics[width=\unitlength,page=2]{volume_manifold.pdf}}%
    \put(0.76878447,0.0038777){\makebox(0,0)[lt]{\lineheight{1.25}\smash{\begin{tabular}[t]{l}$\mathcal{M}$\end{tabular}}}}%
    \put(0,0){\includegraphics[width=\unitlength,page=3]{volume_manifold.pdf}}%
    \put(-0.00153327,0.239376){\makebox(0,0)[lt]{\lineheight{1.25}\smash{\begin{tabular}[t]{l}$V_{G}(x)$\end{tabular}}}}%
    \put(0,0){\includegraphics[width=\unitlength,page=4]{volume_manifold.pdf}}%
    \put(0.19179798,0.03159986){\color[rgb]{1,0,0}\makebox(0,0)[lt]{\lineheight{1.25}\smash{\begin{tabular}[t]{l}$\gamma$\end{tabular}}}}%
    \put(0.83369302,0.05788909){\color[rgb]{1,0,0}\makebox(0,0)[lt]{\lineheight{1.25}\smash{\begin{tabular}[t]{l}$f(\gamma)$\end{tabular}}}}%
  \end{picture}%
\endgroup%

    \caption{Once the low dimensional manifold is equipped with a metric that captures the inherent structure of the high dimensional manifold, we can compute a geodesic $\gamma$, by minimising the energy functional between two points. The geodesic $f(\gamma)$ will be the shortest path between two points on the manifold $\mathcal{M}$. The volume measure $V_G$ can also be used to integrate functions over regions of the manifold, as we would do in the Euclidean space. It can also be linked to the density of the data: if the data points are uniformly distributed over the high-dimensional manifold, in the low-dimensional manifold, a low volume would correspond to a high density of data. It is a useful way to give more information about the distribution of the data.}
\end{figure}
\FloatBarrier

\subsection{Finsler manifolds} \label{subsection:finslermanifold}
Finsler geometry is often described as an extension of Riemannian geometry, since the metric is defined in a more general way, lifting the quadratic constraint. In particular, the norm of a Riemmanian metric is a Finsler metric, but the converse is not true.

\begin{definition}
Let $F:\mathcal{TM} \rightarrow \mathbb{R}_{+}$ be a continuous non-negative function defined on the tangent bundle $\mathcal{TM}$ of a differentiable manifold $M$. We say that $F$ is a \textbf{Finsler metric} if, for each point $x$ of $\mathcal{M}$ and $v$ on ${\mathcal T_x M}$, we have:
\begin{enumerate}
  \item Positive homogeneity: $\forall \lambda \in \mathbb{R}_{+}$, $F(\lambda v) = \lambda F(v)$.
  \item Smoothness: $F$ is a $C^{\infty}$ function on the slit tangent bundle $\mathcal{TM}\setminus{\{0\}}$.
  \item Strong convexity criterion: the Hessian matrix $g_{ij}(v) = \frac{1}{2} \frac{\partial^2 F^2}{\partial v^i v^j}(v)$ is positive definite for non-zero $v$.
\end{enumerate}
A differentiable manifold $\mathcal{M}$ equipped with a Finsler metric is called a \textbf{Finsler manifold}.
\end{definition}

Here, it is worth noting that, for a given point in the manifold, the Finsler metric is defined with only one vector in the tangent space, while the Riemannian metric is defined with two vectors. Moreover, from the previous definition, we can deduce that the metric is:
\begin{enumerate}
  \item Positive definite: for all $x \in \mathcal{M}$ and $v \in \mathcal T_x M$, $F(v) \geq 0$ and $F(v) = 0$ if and only if $v=0$.
  \item Subadditive: $F(v+w) \leq F(v) + F(w)$ for all $x \in \mathcal{M}$ and $v,w \in \mathcal T_x M$.
\end{enumerate}
We say that $F$ is a Minkowski norm on each tangent space ${\mathcal T_x M}$. Furthermore, if $F$ satisfies the reversibility property: $F(v) = F(-v)$, it defines a norm on ${\mathcal T_x M}$ in the usual sense.

Similarly to Riemannian geometry, lengths, energies and volumes can be defined directly from the Finsler metric:
\begin{definition}
We consider a curve $\gamma$ and its derivative $\dot{\gamma}$ on a Finsler manifold $\mathcal{M}$ equipped with the metric $F$. We define the \textbf{length of the curve} as follows:
\[L_F(\gamma) = \int F(\dot{\gamma}(t)) dt.\] 
\end{definition}

\begin{figure}[htbp]
    \centering
    \def\svgscale{0.7}%% Creator: Inkscape 1.0.2 (e86c8708, 2021-01-15), www.inkscape.org
%% PDF/EPS/PS + LaTeX output extension by Johan Engelen, 2010
%% Accompanies image file 'finsler_manifold.pdf' (pdf, eps, ps)
%%
%% To include the image in your LaTeX document, write
%%   \input{<filename>.pdf_tex}
%%  instead of
%%   \includegraphics{<filename>.pdf}
%% To scale the image, write
%%   \def\svgwidth{<desired width>}
%%   \input{<filename>.pdf_tex}
%%  instead of
%%   \includegraphics[width=<desired width>]{<filename>.pdf}
%%
%% Images with a different path to the parent latex file can
%% be accessed with the `import' package (which may need to be
%% installed) using
%%   \usepackage{import}
%% in the preamble, and then including the image with
%%   \import{<path to file>}{<filename>.pdf_tex}
%% Alternatively, one can specify
%%   \graphicspath{{<path to file>/}}
%% 
%% For more information, please see info/svg-inkscape on CTAN:
%%   http://tug.ctan.org/tex-archive/info/svg-inkscape
%%
\begingroup%
  \makeatletter%
  \providecommand\color[2][]{%
    \errmessage{(Inkscape) Color is used for the text in Inkscape, but the package 'color.sty' is not loaded}%
    \renewcommand\color[2][]{}%
  }%
  \providecommand\transparent[1]{%
    \errmessage{(Inkscape) Transparency is used (non-zero) for the text in Inkscape, but the package 'transparent.sty' is not loaded}%
    \renewcommand\transparent[1]{}%
  }%
  \providecommand\rotatebox[2]{#2}%
  \newcommand*\fsize{\dimexpr\f@size pt\relax}%
  \newcommand*\lineheight[1]{\fontsize{\fsize}{#1\fsize}\selectfont}%
  \ifx\svgwidth\undefined%
    \setlength{\unitlength}{481.88961975bp}%
    \ifx\svgscale\undefined%
      \relax%
    \else%
      \setlength{\unitlength}{\unitlength * \real{\svgscale}}%
    \fi%
  \else%
    \setlength{\unitlength}{\svgwidth}%
  \fi%
  \global\let\svgwidth\undefined%
  \global\let\svgscale\undefined%
  \makeatother%
  \begin{picture}(1,0.29411773)%
    \lineheight{1}%
    \setlength\tabcolsep{0pt}%
    \put(0,0){\includegraphics[width=\unitlength,page=1]{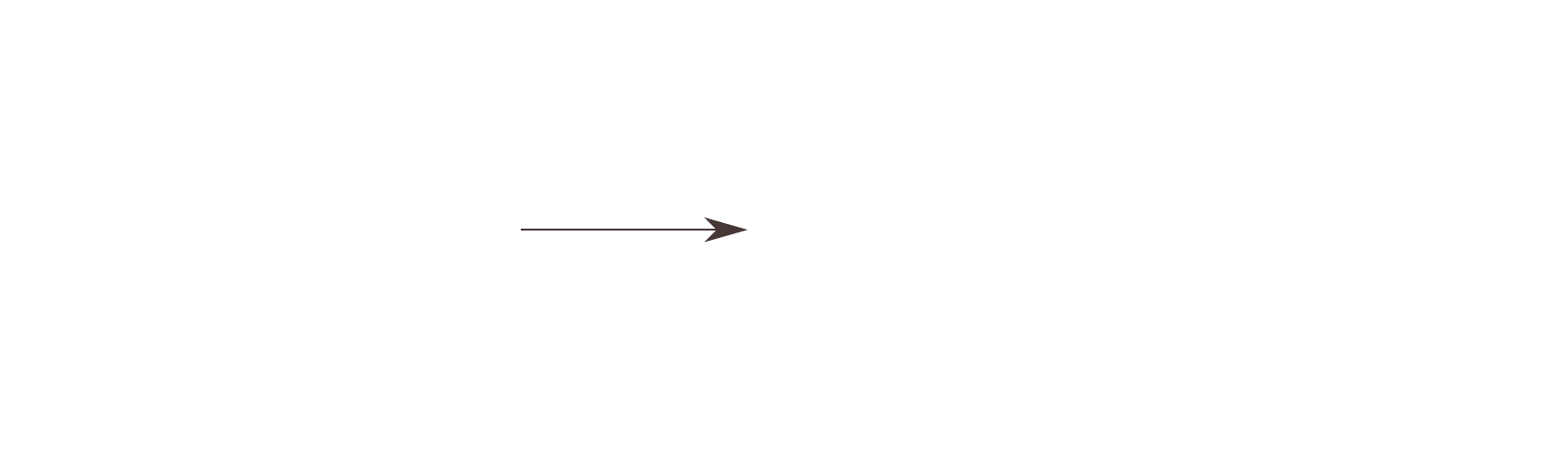}}%
    \put(0.39169362,0.16988129){\makebox(0,0)[lt]{\lineheight{1.25}\smash{\begin{tabular}[t]{l}$f$\end{tabular}}}}%
    \put(0,0){\includegraphics[width=\unitlength,page=2]{finsler_manifold.pdf}}%
    \put(0.13106237,0.09552292){\makebox(0,0)[lt]{\lineheight{1.25}\smash{\begin{tabular}[t]{l}$u$\end{tabular}}}}%
    \put(0.0474289,0.2096414){\color[rgb]{0,0.4,0.50196078}\makebox(0,0)[lt]{\lineheight{1.25}\smash{\begin{tabular}[t]{l}$F_{x}(u) = 1$\end{tabular}}}}%
    \put(0,0){\includegraphics[width=\unitlength,page=3]{finsler_manifold.pdf}}%
    \put(0.76386382,0.00395779){\makebox(0,0)[lt]{\lineheight{1.25}\smash{\begin{tabular}[t]{l}$\mathcal{M}$\end{tabular}}}}%
    \put(0.57438695,0.21966717){\makebox(0,0)[lt]{\lineheight{1.25}\smash{\begin{tabular}[t]{l}$\mathcal{T}_x \mathcal{M}$\end{tabular}}}}%
    \put(0,0){\includegraphics[width=\unitlength,page=4]{finsler_manifold.pdf}}%
  \end{picture}%
\endgroup%

    \caption{$f$ is an immersion that maps a low dimensional manifold to a high dimensional manifold $\mathcal{M}$. On $\mathcal{M}$, a tangent plane $\mathcal{T}_x \mathcal{M}$ is draw at $x$. Compared to the Riemannian manifold, the Finsler indicatrix, which represents the all the vectors $u \in \mathcal{T}_x \mathcal{M}$ such that $F_x(u)=1$, is not necessarily an ellipse. It can be asymmetric if the metric is asymmetric itself. It is always convex.}
\end{figure}
\FloatBarrier

\begin{definition}
The \textbf{curve energy} is defined as: $E_F(\gamma) = \int F(\dot{\gamma}(t))^2 dt.$
\end{definition}
Not only are the definitions strikingly similar, they also share the same properties. The curve length is also invariant under reparametrisation, and upper bounded by the curve energy. Computing geodesics on a manifold is reduced to a variational optimisation problem. These propositions are proved in detail in Lemmas \ref{lemma:finslerreparametrisation} and \ref{lemma:minimizationenergy}, in the appendix.

In Riemannian geometry, the volume measure defined by the metric is unique. In Finsler geometry, different definitions of the volume exist, and they all coincide with the Riemannian volume element when the metric is Riemannian. The most common choices of volume forms are the Busemann-Hausdorff measure and the Holmes-Thompson measure. According  \citet{wu:2011}, depending on the Finsler metric and the topological manifold, some choices seem more legitimate than others. In this paper, we decided to only focus on the Busemann-Hausdorff volume, as its definition is the most commonly used and leads to easier derivations. 
% In this paper, because our Finsler metric is reversible and we don't restrain ourselves to totally geodesic manifolds, we decided to retrain ourself to the Busemann-Hausdorff definition. NOT TRUE?
We will later show that in high dimensions, our Finsler metric converges to a Riemannian metric, and thus, the results obtained for the Busemann-Hausdorff volume measure are also valid for the Holmes-Thomson volume measure.

\begin{definition}
For a given point $x$ on the manifold, we define the \textbf{Finsler indicatrix} as the set of vectors in the tangent space such that the Finsler metric is equal to one: $\{v \in {\mathcal T_x M} | F(v) = 1\})$. We denote the Euclidean unit ball in $\RR^d$ by $\BB^{d}(1)$ and for measurable subsets $S \subseteq \RR^d$ we use $\vol(S)$ to denote the standard Eulcidean volume of $S$. In local coordinates ($e^1, \cdots, e^d$) on a Finsler manifold $\mathcal{M}$, the \textbf{Busemann-Hausdorff volume} form is defined as $dV_F = V_F(x) e^1 \wedge \cdots \wedge e^d $, with:
\[V_F(x) = \frac{\vol(\BB^d(1))}{\vol(\{v \in {\mathcal T_x M} | F(v) < 1\})}.\]
\end{definition}

We can interpret the volume as the ratio between the euclidean ball, and a convex ball whose radius is defined as a unit Finsler metric. If the Finsler metric is replaced by a Riemannian metric, the volume of the indicatrix will be an ellipsoid whose semi-axis are equal to the inverse of the squareroot of the metric's eigenvalues. The Finsler volume then reduces to the definition of the Riemannian volume. 
% {\color{blue} I can add the proof if it's useful to have a better understanding of the definition.}

\subsection{Random manifolds} \label{subsection:randommanifold}
So far, we have only considered deterministic data points lying on a manifold. If we consider our data to be random variables, we will need to define the associated random metric and manifold. 

% \FloatBarrier
% \begin{figure}[htbp]
%     \centering
%     \includegraphics[width=0.6\textwidth]{figures/primer_geometry/illustration_random_manifolds.png}
%     \caption{In this figure, the map between the latent space and the observational space is stochastic. Each sampling will lead to different data points, and so different manifolds (blue, red and purple). We need to take this randomness into account when we want to compute length, energy or volume. For computing a geodesic, we can either average all the geodesic obtained, or compute the geodesic with the expected metric. (see next section)}
%     % \label{fig:}
% \end{figure}

% \begin{figure}[htbp]
%     \centering
%     \includegraphics[width=0.5\textwidth]{figures/primer_geometry/test.png}
%     \caption{Something about stochastic manifolds}
% \end{figure}
% \FloatBarrier

\begin{figure}[htbp]
    \centering
    \def\svgscale{0.7}%% Creator: Inkscape 1.0.2 (e86c8708, 2021-01-15), www.inkscape.org
%% PDF/EPS/PS + LaTeX output extension by Johan Engelen, 2010
%% Accompanies image file 'random_manifold2.pdf' (pdf, eps, ps)
%%
%% To include the image in your LaTeX document, write
%%   \input{<filename>.pdf_tex}
%%  instead of
%%   \includegraphics{<filename>.pdf}
%% To scale the image, write
%%   \def\svgwidth{<desired width>}
%%   \input{<filename>.pdf_tex}
%%  instead of
%%   \includegraphics[width=<desired width>]{<filename>.pdf}
%%
%% Images with a different path to the parent latex file can
%% be accessed with the `import' package (which may need to be
%% installed) using
%%   \usepackage{import}
%% in the preamble, and then including the image with
%%   \import{<path to file>}{<filename>.pdf_tex}
%% Alternatively, one can specify
%%   \graphicspath{{<path to file>/}}
%% 
%% For more information, please see info/svg-inkscape on CTAN:
%%   http://tug.ctan.org/tex-archive/info/svg-inkscape
%%
\begingroup%
  \makeatletter%
  \providecommand\color[2][]{%
    \errmessage{(Inkscape) Color is used for the text in Inkscape, but the package 'color.sty' is not loaded}%
    \renewcommand\color[2][]{}%
  }%
  \providecommand\transparent[1]{%
    \errmessage{(Inkscape) Transparency is used (non-zero) for the text in Inkscape, but the package 'transparent.sty' is not loaded}%
    \renewcommand\transparent[1]{}%
  }%
  \providecommand\rotatebox[2]{#2}%
  \newcommand*\fsize{\dimexpr\f@size pt\relax}%
  \newcommand*\lineheight[1]{\fontsize{\fsize}{#1\fsize}\selectfont}%
  \ifx\svgwidth\undefined%
    \setlength{\unitlength}{483.26597995bp}%
    \ifx\svgscale\undefined%
      \relax%
    \else%
      \setlength{\unitlength}{\unitlength * \real{\svgscale}}%
    \fi%
  \else%
    \setlength{\unitlength}{\svgwidth}%
  \fi%
  \global\let\svgwidth\undefined%
  \global\let\svgscale\undefined%
  \makeatother%
  \begin{picture}(1,0.29328005)%
    \lineheight{1}%
    \setlength\tabcolsep{0pt}%
    \put(0,0){\includegraphics[width=\unitlength,page=1]{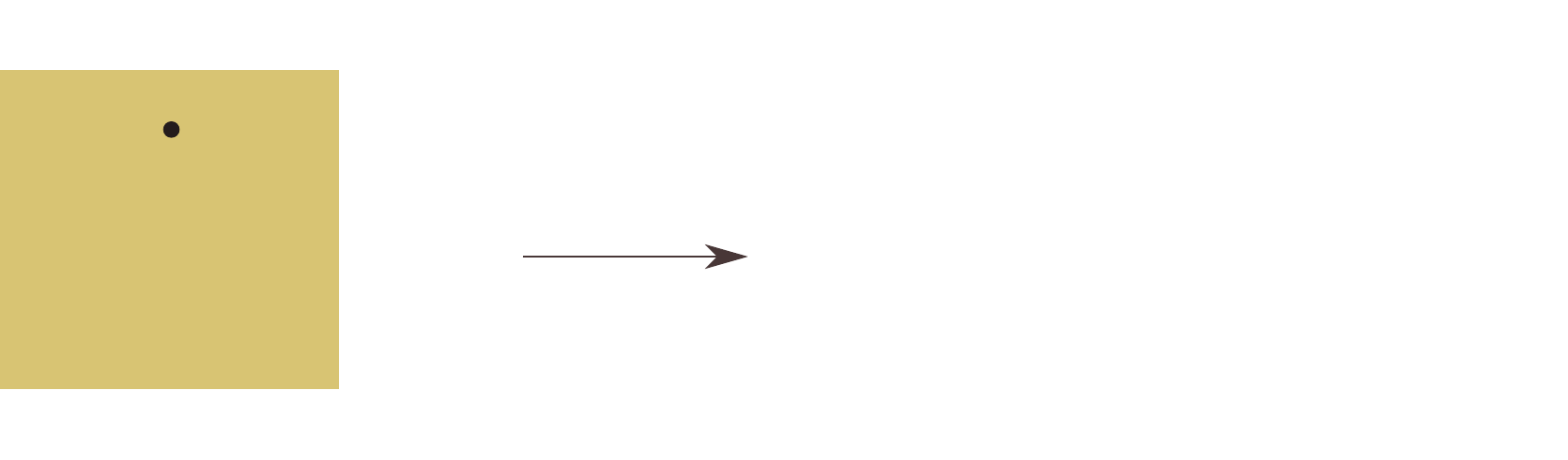}}%
    \put(0.79626998,0.00401585){\makebox(0,0)[lt]{\lineheight{1.25}\smash{\begin{tabular}[t]{l}$\mathcal{M}$\end{tabular}}}}%
    \put(0,0){\includegraphics[width=\unitlength,page=2]{random_manifold2.pdf}}%
    \put(0.17548143,0.10012233){\color[rgb]{0,0,0}\transparent{0.69999999}\makebox(0,0)[lt]{\lineheight{0}\smash{\begin{tabular}[t]{l}$G_1$\end{tabular}}}}%
    \put(0.17337527,0.07325948){\color[rgb]{0,0.26666667,0.33333333}\transparent{0.69999999}\makebox(0,0)[lt]{\lineheight{0}\smash{\begin{tabular}[t]{l}$G_2$\end{tabular}}}}%
    \put(0.174511,0.04926519){\color[rgb]{0.12941176,0.40392157,0.47058824}\transparent{0.69999999}\makebox(0,0)[lt]{\lineheight{0}\smash{\begin{tabular}[t]{l}$G_3$\end{tabular}}}}%
    \put(0.33186249,0.14815735){\makebox(0,0)[lt]{\lineheight{1.25}\smash{\begin{tabular}[t]{l}$\{f_t\} = \{f_1, f_2, f_3, \cdots\}$\end{tabular}}}}%
  \end{picture}%
\endgroup%

    \caption{Usually the immersion $f$ would be deterministic. In the case of most generative models, where $f$ is described by a GP-LVM, or the decoder of a VAE, the immersion is stochastic. The pullback metric is stochastic de facto.}
\end{figure}
\FloatBarrier

As said previously, if we have a function $f: \RR^q \rightarrow \RR^D$ that parametrises a manifold, then we can construct a Riemannian metric $G=J_{f}^{\top}J_f$, with $J_f$ the Jacobian of the function $f$. In the previous cases, we assumed $f$ to be a deterministic function, and so is the metric. We construct a stochastic Riemannian metric in the same way, with $f$ being a stochastic process. A stochastic process $f:\RR^q \rightarrow \RR^D$ is a random map in the sense that samples of the process are maps from $\RR^q$ to $\RR^D$ (the so-called sample paths of the process).

\begin{definition}
A stochastic process $f: \RR^q \rightarrow \RR^D$ is smooth if the sample paths of $f$ are smooth. We call a smooth process f a \textbf{stochastic immersion} if the Jacobian matrix of its sample paths has full rank everywhere. We can then define the \textbf{stochastic Riemannian metric} $G = J_{f}^{\top}J_f$.
\end{definition}

The terms \textit{stochastic} and \textit{random} are used interchangeably. The definition of the stochastic immersion is fairly important, as it means that its Jacobian is full rank. Since the Jacobian is full rank, the random metric $G$ is positive definite, a necessary condition to define a Riemannian metric. Another definition of a stochastic Riemannian metric would be the following:
\begin{definition}
A \textbf{stochastic Riemannian metric} on $\RR^{q}$ is a matrix-valued random field on $\RR^{q}$ whose sample paths are Riemannian metrics. A \textbf{stochastic manifold} is a differentiable manifold equipped with a stochastic Riemannian metric.
\end{definition}

Any matrice drawn from this stochastic metric would be a proper Riemannian metric. When using the random Riemannian metric on two vectors $u,v \in {\mathcal T_x M}$, $G(u,v)=u^{\top} G v$ is a random variable, but both $u,v$ are deterministic vectors. From this definition, it follows that the length, the energy and the also volume are random variables. 

% \subsection{Summary}

\begin{table}[htbp]
\centering
\begin{tabular}{|l|l|l|} \hline
Object & Riemann & Finsler   \\[0.4cm]
\hline \hline
metric & $g:{\mathcal T_x M} \times {\mathcal T_x M} \rightarrow \mathbb{R}$   & $F: {\mathcal T M} \rightarrow \mathbb{R_{+}}$ \\[0.4cm]
length structure  & $L_{G}(\gamma) = \int \sqrt{g_t(\dot{\gamma}(t),\dot{\gamma}(t))} \,dt$ & $L_F(\gamma) = \int F(\dot{\gamma}(t)) \,dt$ \\[0.4cm]
energy structure  & $E_G(\gamma) = \int g_t(\dot{\gamma}(t),\dot{\gamma}(t)) \,dt$ & $E_F(\gamma) = \int F(\dot{\gamma}(t))^2 \,dt$ \\[0.4cm]
volume element & $V_G(x) = \sqrt{|\det{G}|}$ &$V_F(x)= \vol(\BB^n(1))/ \vol(\{v \in {\mathcal T_x M} | F(x,v) < 1\})$ \\
{} & {} & Busemann-Hausdorff volume measure \\[0.4cm]
\cline{1-3}
\end{tabular}
\caption{Comparison of Riemannian and Finsler metrics.}
\end{table}

\section{Proofs}
\label{section:Proofs}
One of the main challenges of this paper is to find coherent notations while respecting the tradition of two geometric fields. In Riemannian geometry, we principally use a metric, noted $g_p: {\mathcal T_p \mathcal{M}} \times {\mathcal T_p \mathcal{M}} \to \RR_+$, that is defined as an inner product and thus can induce a norm, but is not a norm. In Finsler geometry, we call intercheangeably Finsler function, Finsler metric or Finsler norm, the norm traditionally noted $F: \mathcal{TM} \to \RR_+$, with $F_p(u) \coloneqq F(p,u)$ defined at a point $p\in\mathcal{M}$ for a vector $u\in {\mathcal T_p \mathcal{M}}$. We will assume that all our metric are always defined for a specific point $z$ (or $p$) on our manifold $\mathcal{Z}$ (or $\mathcal{M}$), and so we will just drop this index. The following notations will be used: 

\begin{tabular}{cp{0.8\textwidth}}
  Stochastic pullback metric tensor & $G = J_f^{\top} J_f$ \\
  Stochastic pullback metric & $\Tilde{g}: (u,v) \to u^{\top}G u$ \\
  Expected Riemannian metric & $g: (u,v) \to u^{\top}\EE[G] v$ \\
  Stochastic pullback induced norm & $\norm{\cdot}_G: u \to \sqrt{u^{\top}G u}$ \\
  Expected Riemannian induced norm & $\norm{\cdot}_R: u \to \sqrt{u^{\top}\EE[G] u} \coloneqq \sqrt{g(u,u)}$ \\
  Finsler metric & $\norm{\cdot}_F: u \to \EE[\sqrt{u^{\top}G u}] \coloneqq F(u)$ \\
\end{tabular}

\subsection{Finslerian geometry of the expected length} 
\label{appendix:proof:finslergeometry}
In this section, we will always let $f: \RR^q \rightarrow \RR^D$ be a stochastic immersion, $J_f$ its Jacobian, and $G = J_f^{\top} J_f$ a metric tensor. We will first prove that the function $ F: {\mathcal T M} \rightarrow \RR: v \rightarrow \EE \left[\sqrt{v^{\top} G v}\right]$ is a Finsler metric. Then, for the specific case where $J_f$ follows a non-central normal distribution, the Finsler metric $F$ defined as the expected length follows a non-central Nakagami distribution and can be expressed in closed form. \\

To prove that the function $F$ is indeed a Finsler metric, we will need to verify the criteria above, among them the strong convexity criterion is less trivial to prove than the others. It will be detailed in Lemma \ref{lemma:strongconvexity}. Strong convexity means that the Hessian matrix $\frac{1}{2}\Hess(F(v)^2) = \frac{1}{2} \frac{\partial^2 F^2}{\partial v^i v^j}(v)$ is strictly positive definite for non-negative $v$. This matrix, when $F$ is a Finsler function, is also called the fundamental form and plays an important role in Finsler geometry. To prove the strong convexity criterion, we will need the full expression of the fundamental form, detailed in Lemma \ref{lemma:fundamentalform}.

\begin{lemma} \label{lemma:fundamentalform}
The Hessian matrix  $\frac{1}{2}\Hess(F(v)^2)$ of the function $F(v) = \EE \left[\sqrt{v^{\top} G v}\right]$ is given by
\[ \frac{1}{2}\Hess(F(v)^2)  = \EE\left[(v^{\top} G v)^{\frac{1}{2}}\right]\EE\left[(v^{\top} G v)^{-\frac{1}{2}} G -(v^{\top} G v)^{-\frac{3}{2}} G vv^{\top} G\right]+\EE\left[(v^{\top} G v)^{-\frac{1}{2}} G\right]^2 vv^{\top} .\]
\end{lemma}

\begin{proof}
Let $G$ be a random positive definite symmetric matrix and define $g: \RR^{q} \rightarrow \RR: v \mapsto \sqrt{v^{\top} G v}$, where $v$ is considered a column vector. We would like to know the different derivatives of $g$ with respect to $v$. We name by default $J_g$ and $H_g$, its Jacobian and Hessian matrix. Using the chain rule, we have: $J_g = (v^{\top} G v)^{-\frac{1}{2}} v^{\top}G$ and $H_g =  (v^{\top} G v)^{-\frac{1}{2}} G - (v^{\top} G v)^{-\frac{3}{2}} (G vv^{\top} G)$. 

For the rest of the proof, we need to show that derivatives and expectation values commute.
% : $\frac{\partial \EE[g]}{\partial v_i} = \EE[\frac{\partial g}{\partial v_i}]$. 
% The derivatives $\frac{\partial g}{\partial v}$ being always upper bounded by $v^{\top} G$ and $\EE[v^{\top} G] < \infty$, using the theorem of dominated convergence, we have that the expectation values and derivatives commute. In other words, computing the Jacobian of the expectation of the function $g$ is the same as computing the expectation of the Jacobian $J_g$, and similarly for the Hessian $H_g$. 

Using the Fubini theorem, we can show that tha derivatives and the expectation values commute.

For $F: \RR^{q} \rightarrow \RR: v \mapsto \EE[\sqrt{v^{\top} G v}]$,  
\[\Hess(F) = \EE[H_g] = \EE\left[(v^{\top} G v)^{-\frac{1}{2}} G -(v^{\top} G v)^{-\frac{3}{2}} G vv^{\top} G\right]\]
\[\nabla F = \EE[J_g] = \EE[(v^{\top} G v)^{-\frac{1}{2}} G v]. \] 

We now consider the function $h: \RR^{q} \rightarrow \RR: v\mapsto \EE[\sqrt{v^{\top} G v}]^2 = F(v)^2$. Using the chain rule and changing the order of expectation and derivatives, we have its Hessian \[H_h = 2F \cdot \Hess[F] + 2\nabla F^{\top} \nabla F = 2 \EE[g] \EE[H_g] + 2 \EE[J_g]^{\top}\EE[J_g].\]

Finally, replacing $J_g$ and $H_g$ previously obtained in this expression, we conclude: \[ \frac{1}{2} H_h(x,v) = \EE\left[(v^{\top} G v)^{\frac{1}{2}}\right]\EE\left[(v^{\top} G v)^{-\frac{1}{2}} G -(v^{\top} G v)^{-\frac{3}{2}} G vv^{\top} G\right]+\EE\left[(v^{\top} G v)^{-\frac{1}{2}} G\right]^2 vv^{\top}. \]
\end{proof}

\begin{remark}
Before going further, it's important to note that $G = J_f^{\top} J_f$ is a random matrix that is positive definite: it is symmetric by definition and has full rank. The later statement is justified by the assumption that the stochastic process $f:\RR^{q} \rightarrow \RR^{D}$ is an immersion, then $J_f$ is full rank.
\end{remark}

\begin{lemma} \label{lemma:homogeneouscontinuity}
The function $F(v) = \EE \left[\sqrt{v^{\top} G v}\right]$ is:
    \begin{enumerate}
      \item positive homogeneous: $\forall \lambda \in \mathbb{R}_{+}$, $F( \lambda v) = \lambda F(v)$
      \item smooth: $F(v)$ is a $C^{\infty}$ function on the slit tangent bundle $\mathcal{TM}\setminus{\{0\}}$
    \end{enumerate}
\end{lemma}

\begin{proof}
1) Let $\lambda \in \RR$, then we have: $F(\lambda v) = \EE \left[\sqrt{\lambda^2 v^{\top} G v}\right] = |\lambda|\left[\sqrt{v^{\top} G v}\right]$. 

2) The multivariate function: $\RR^q\backslash\{0\} \rightarrow \RR_{+}^{*}: v \rightarrow v^{\top} G v$ is $C^{\infty}$ and strictly positive, since $G = J_f^{\top} J_f$ is positive definite. The function $\RR_{+}^{*} \rightarrow \RR_{+}^{*}: x \rightarrow \sqrt{x}$ is also $C^{\infty}$. Finally, $\RR_{+}^{*} \rightarrow \RR_{+}^{*}: x \rightarrow \EE[x]$ is by definition differentiable. By composition, $F(v)$ is a $C^{\infty}$ function on the slit tangent bundle $\mathcal{TM}\setminus{\{0\}}$.
\end{proof}

\begin{lemma} \label{lemma:strongconvexity}
The function $F(v) = \EE \left[\sqrt{v^{\top} G v}\right]$ satisfies the strong convexity criterion.
\end{lemma}

\begin{proof}
Proving that F satisfies the strong convexity criterion is equivalent to show that the Hessian matrix $H = \frac{1}{2}\Hess(F(v)^2)$ is strictly positive definite. Thus, we need to prove that $\forall w \in \RR^{q} \backslash \{0\}, w^{\top} H w > 0$. According to Lemma \ref{lemma:fundamentalform}, because the expectation is a positive function, it's straightforward to see that $\forall w \in \RR^{q} \backslash \{0\}, w^{\top} H w \geq 0$. The tricky part of this proof is to show that $w^{\top} H w > 0$. This can be obtained if one of the terms ($F \cdot \Hess(F)$ or $\nabla F^{\top} \nabla F$) is strictly positive. 

First, let's decompose $H$ as the sum of matrices: $H=F \Hess(F) + \nabla F^{\top} \nabla F$ (Lemma \ref{lemma:fundamentalform}), with: 
\[F \cdot \Hess(F) = \EE \left[(v^{\top} G v)^{\frac{1}{2}}\right]\EE\left[(v^{\top} G v)^{-\frac{3}{2}} \left((v^{\top} G v)G - Gv (Gv)^{\top}\right)\right],\] \[\nabla F^{\top} \nabla F = \EE\left[(v^{\top} G v)^{-\frac{1}{2}} G\right]^2 vv^{\top}.\] We will study two cases: when $w \in \Span(v)$, and when $w \notin \Span(v)$. We will always assume that $v \neq 0$, and so by definition: $F(v)>0$.

Let $w \in \Span(v)$. We will show that $w^{\top} \nabla F^{\top} \nabla F w>0$. We have $w=\alpha v, \alpha \in \RR$. Because F is 1-homogeneous and using Euler theorem, we have: $\nabla F(v) v = F(v)$. Then $(\alpha v)^{\top} \nabla F^{\top} \nabla F (\alpha v) = \alpha^2 F^2$, and $\alpha^2 F(v)^2 > 0$.

Let $w \notin \Span(v)$. F being a scalar function, we have: $w^{\top} F \Hess[F] w = F w^{\top} \Hess[F] w$. We would like to show that: $w^{\top} \Hess[F] w > 0$. The strategy is the following: if we prove that the kernel of $\Hess[F]$ is equal to the $\Span(v)$, then $w \notin \Span(v)$ is equivalent to say that $w \notin \ker(\Hess[F])$ and we can conclude that: $w^{\top} \Hess[F] w > 0$. Let's prove $\Span(v) \in \ker(\Hess(F))$.  We know that $\Hess(F)v = 0$, since F is 1-homogeneous, so we have $\Span(v) \in \ker(\Hess(F))$. To obtain the equality, we just need to prove that the dimension of the kernel is equal to 1. Let $z \in \Span(v^{\top}G)^{\top}$, which is $(Gv)^{T}z=0$. We have $\dim(\Span(v^{\top}M))=1$, and thus: $\dim(\Span(v^{\top}G)^{\top})=q-1$. Furthermore, $z^{\top} \Hess[F] z = z^{\top}\EE\left[M (v^{\top} M v)^{-\frac{1}{2}}\right] z > 0$, so we can deduce that $\dim(\im(\Hess[F]))=q-1$. Using the Rank-Nullity theorem, we conclude that $\dim(\ker(\Hess(F))) = q- \dim(\im(\Hess[F])) = 1$, which concludes the proof.

In conclusion, $\forall w \in \RR^{q} \backslash \{0\}, w^{\top} \frac{1}{2}\Hess(F(v)^2) w > 0$. The function $F$ satisfies the strong convexity criterion.
\end{proof}

\propDefinitionFinslerOne*
\begin{proof}
Let's define F as a Riemannian metric: $F: \RR^{q} \times \RR^{q} \rightarrow \RR: (v_1, v_2) \rightarrow \EE\left[\sqrt{v_1^{\top} G v_2}\right]$. If $F$ were a Riemannian metric, then it would be bilinear, which is clearly not the case. Thus, $F$ is not a Riemannian metric. According to Lemma \ref{lemma:homogeneouscontinuity} and Lemma \ref{lemma:homogeneouscontinuity}, $F$ is a Finsler metric.
\end{proof}

\propDefinitionFinslerTwo*

\begin{proof}
The objective of the proof is to show that, if the Jacobian $J_f$ follows a non-central normal distribution, then, $\forall v \in \RR^q$, the expectation $\EE[v^{\top} J_f^{\top} J_f v]$ will follow a non-central Nakagami distribution. This is a particular case of the derivation of moments of non-central Wishart distributions, previously shown and studied by \cite{muirhead:1984, hauberg:2018}. \\

By hypothesis, $J_f$ follows a non-central normal distribution: $J_f \sim \mathcal{N}(\EE[J], I_D \otimes\Sigma)$. Then, $G = J_f^{\top} J_f$ follows a non-central Wishart distribution: $G \sim \mathcal{W}_d(D, \Sigma, \Sigma^{-1} \EE[J]^{\top} \EE[J])$. According to \cite[Theorem~10.3.5.]{muirhead:1984}, $v^{\top} G v$ will also follow a non-central Wishart distribution: $v^{\top} G v \sim \mathcal{W}_1(D, v^{\top} \Sigma v, \omega)$,
with: $\omega= (v^{\top} \Sigma v)^{-1}(v^{\top} \EE[J]^{\top} \EE[J] v) $. \\

To compute $\EE[\sqrt{v^{\top} G v}]$, we shall look at the derivation of moments. \cite[Theorem~10.3.7.]{muirhead:1984} states that: if $X \sim \mathcal{W}_q(D, \Sigma, \Omega')$, with $q \leq D$, then
$\EE[(\det(X))^k] = (\det{\Sigma})^k 2^{qk} \frac{\Gamma_q(\frac{D}{2} + k)}{\Gamma_q(\frac{D}{2})}\kummer(-k, \frac{D}{2}, -\frac{1}{2}\Omega')$. We directly apply the theorem to our case, knowing that $v^{\top} G v$ is a scalar term, so $\det(v^{\top} G v)= v^{\top} G v$, $q=1$, and $k=\frac{1}{2}$: 

\[\norm{v}_F \coloneqq \EE[\sqrt{v^{\top} G v}] = \sqrt{2} \sqrt{v^{\top} \Sigma v}  \frac{\Gamma(\frac{D}{2} + \frac{1}{2})}{\Gamma(\frac{D}{2})}\kummer(-\frac{1}{2}, \frac{D}{2}, -\frac{1}{2}\omega) \]
\end{proof}

\begin{lemma} \label{lemma:finslerreparametrisation}
The length of a curve using a Finsler metric is invariant by reparametrisation.
\end{lemma}

\begin{proof}
The proof is similar to the one obtained on a Riemannian manifold (\cite{lee:2013}, Proposition 13.25), where we make use of the homogeneity property of the Finsler metric. 

Let ($\mathcal{M}, F$) be a Finsler manifold and $\gamma: [a,b] \to \mathcal{M}$ a piecewise smooth curve segment. We call $\tilde{\gamma}$ a reparametrisation of $\gamma$, such that $\tilde{\gamma} = \gamma \circ \phi$ with $\phi: [c,d] \to [a,b]$ a diffeomorphism. We want to show that $L_F(\gamma)=L_F(\tilde{\gamma})$.

\begin{equation*}
\begin{aligned}
L_F(\tilde{\gamma}) = &  \int_{c}^{d} F(\dot{\tilde{\gamma}}(t)) \,dt = \int_{c}^{d} F(\frac{d}{dt}(\gamma \circ \phi (t))) \,dt \\ 
                            = &  \int_{\phi^{-1}(a)}^{\phi^{-1}(b)} |\dot{\phi} (t)| F( \dot{\gamma} \circ \phi (t)) \,dt = \int_{a}^{b} F(\dot{\gamma}(t)) \,dt = L_F(\gamma)
\end{aligned}
\end{equation*}

\end{proof}

\begin{lemma} \label{lemma:minimizationenergy}
If a curve globally minimizes its energy on a Finsler manifold, then it also globally minimizes its length and the Finsler function $F$ of the velocity vector along the curve is constant.
\end{lemma}

\begin{proof}
The curve energy and the curve length are defined as:
  $E_F(\gamma) = \int_0^1 F^2(\dot{\gamma}(t)) dt$ and $L_F(\gamma) = \int_0^1 F(\dot{\gamma}(t)) dt$, with $\gamma:[0,1] \to \RR^d$.
Let's define $f$ and $g$ two real-valued functions such that: $f: \RR \to \RR: t \mapsto F( \dot{\gamma}(t))$ and
 $g: \RR \to \RR: t \mapsto 1$. Applying Cauchy-Schwartz inequality, we directly obtain:
\[\left( \int_0^1 F(\dot{\gamma}(t))dt \right)^{2} \leq \int_0^1 F(\dot{\gamma}(t))^{2}dt \cdot \int_0^1 1^{2}dt, \quad \text{which means:} \quad L_F(\gamma)^{2} \leq E_F(\gamma).\]
  The equality is obtained exactly when the functions $f$ and $g$ are proportional, hence, when the Finsler function is constant.
\end{proof}

\subsection{Comparison of Riemannian and Finsler metrics}
We have defined both a Riemannian ($g:(v_1, v_2) \rightarrow v_1^{\top}\EE[G]v_2$) and a Finsler ($F:(x,v) \rightarrow \EE[\sqrt{v^{\top} G v}]$) metric, in the hope to compute the average length between two points on a random manifold created by the random field $f$: $G = J_f^{\top} J_f$. The main idea of this section is to better compare those two metrics and in what extend they differ in terms of length, energy and volume. From now on, $f: \RR^q \rightarrow \RR^D$ will always be defined as a stochastic non-central gaussian process. Its Jacobian $J_f$ also follows a non-central gaussian distribution, $G = J_f^{\top} J_f$ a non-central Wishart distribution, and $F:(x,v) = \EE[\sqrt{v^{\top} G v}]$ a non-central Nakagami distribution (Proposition \ref{proposition:noncentralNakagami}). The Finsler metric can be written in closed form.\\

In section \ref{appendix:absolutebounds}, we will see that the Finsler metric is upper and lower bounded by two Riemannian tensors (Proposition \ref{proposition:absolutebounds}), and we can deduce an upper and lower bound for the length, the energy and the volume (Corollary \ref{corollary:absolutebounds}). Then, in section \ref{appendix:relativebounds}, we will show that the relative difference between the Finsler norm and the Riemannian induced norm is always positive and upper bounded a term that is inversely proportional to the number of dimensions $D$ (Proposition \ref{proposition:relativebounds}). Similarly, we will deduce the same for the length, the energy and the volume (Corollary \ref{corollary:relativebounds}). From this last results, we can directly conclude in section \ref{appendix:highdimensions} that both metrics are equal in high dimensions (Corollary \ref{corollary:highdimensionsmetrics}). A possible interpretation is that in high dimensions the data distribution obtained on those manifolds becomes more and more concentrated around the mean, reducing the variance term to zero. The manifold becoming deterministic, both metrics become equal.

\begin{remark}
Most of the following proofs will be a bit technical, as they rely on  the closed form expression of the non-central Nakagami distribution. Once proving the main propositions, obtaining the corollaries is straightforward. While we do not have closed form expression of the indicatrix, we will show that it's a monotoneous function which can upper and lower bounded. 
\end{remark}

\subsubsection{Bounds on the Finsler metric} \label{appendix:absolutebounds}

% \begin{lemma}
% Let $G$ be defined as: $G \sim \mathcal{W}_q(D, \Sigma, \Omega)$ a non-central Wishart distribution. Then:
% \[\EE[G] = D\Sigma + \Sigma \Omega\]. 
% \end{lemma}
% \begin{proof}
% From \cite[p.~442]{muirhead:1984}, the expectation of a non-central Wishart distribution ($G \sim \mathcal{W}_q(D, \Sigma, \Omega)$).
% \end{proof}

\propabsolutebounds*
\begin{proof}
Let's first recall that the Finsler function can be written as:
\[\norm{v}_F \coloneqq F(v) = \sqrt{2} \sqrt{v^{\top} \Sigma v}  \frac{\Gamma(\frac{D}{2} + \frac{1}{2})}{\Gamma(\frac{D}{2})}\cdot \kummer(-\frac{1}{2}, \frac{D}{2}, -\frac{1}{2}\omega).\]

The confluent hypergeometric function is defined as: $ \kummer(a,b,z) = \sum_{k=0}^{\infty} \frac{(a)_{k}}{(b)_{k}} \frac{z^{k}}{k !}, $ with $(a)_k$ and $(b)_k$ being the Pochhammer symbols. Note that, despite their confusing notation, they are defined as rising factorials. By definition, we have: $\frac{(a)_{k}}{(b)_{k}} = \frac{\Gamma(a+k)}{\Gamma(b+k)} \frac{\Gamma(b)}{\Gamma(a)}$. We can use the Kummer transformation to obtain: 

$\kummer(a, b, -z) = e^{-z}\kummer(b-a, b, z)$. Replacing $a=-\frac{1}{2}$, $b=\frac{D}{2}$ and $z=\frac{1}{2}\omega$, we finally get:
\[F(v) = \sqrt{2} \sqrt{v^{\top} \Sigma v} \cdot e^{-z} \sum_{k=0}^{\infty}  \frac{\Gamma(\frac{D}{2}+\frac{1}{2}+k)}{\Gamma(\frac{D}{2} +k)} \frac{z^k}{k!}.\]

1) Let's show that: $\forall v \in {\mathcal T_x M}: \  \sqrt{v^{\top} \alpha  \Sigma v} \leq F(v)$, with $\alpha = 2 \left(\frac{\Gamma(\frac{D}{2}+\frac{1}{2})}{\Gamma (\frac{D}{2})}\right)^2 $. 

The Pochhammer symbole is defined as $(x)_k = x(x+1)\dots (x+k-1)=\frac{\Gamma(x+k)}{\Gamma(x)}$.
For $x \in \RR_{+}^{*}$, we have: $(x)_k \leq (x+\frac{1}{2})_k$. Thus, $\frac{\Gamma(\frac{D}{2}+k)}{\Gamma(\frac{D}{2})} \leq \frac{\Gamma(\frac{D}{2}+\frac{1}{2}+k)}{\Gamma(\frac{D}{2}+\frac{1}{2})}$.
The Gamma function being strictly positive on $\RR_{+}$, we obtain:

\begin{equation*}
\begin{aligned}
\frac{\Gamma(\frac{D}{2}+\frac{1}{2})}{\Gamma(\frac{D}{2})} {} & \leq \frac{\Gamma(\frac{D}{2}+\frac{1}{2}+k)}{\Gamma(\frac{D}{2}+k)} \\ 
\sqrt{2} \sqrt{v^{\top} \Sigma v} \  \frac{\Gamma(\frac{D}{2}+\frac{1}{2})}{\Gamma(\frac{D}{2})} \cdot  e^{-z}  \sum_{k=0}^{\infty}  \frac{z^k}{k!} {} & \leq \sqrt{2} \sqrt{v^{\top} \Sigma v} \cdot e^{-z} \sum_{k=0}^{\infty}  \frac{\Gamma(\frac{D}{2}+\frac{1}{2}+k)}{\Gamma(\frac{D}{2} +k)} \frac{z^k}{k!}  \\
\sqrt{2} \frac{\Gamma(\frac{D}{2}+\frac{1}{2})}{\Gamma(\frac{D}{2})} \sqrt{v^{\top} \Sigma v}  {} & \leq \sqrt{2} \sqrt{v^{\top} \Sigma v} \cdot e^{-z} \sum_{k=0}^{\infty}  \frac{\Gamma(\frac{D}{2}+\frac{1}{2}+k)}{\Gamma(\frac{D}{2} +k)} \frac{z^k}{k!}  \\
\sqrt{v^{\top} \alpha  \Sigma v} {} & \leq  F(v).  \\
\end{aligned}
\end{equation*} 

2) Let's show that: $\forall v \in {\mathcal T_x M}: \ F(v) \leq \sqrt{v^{\top}\EE[G]v}$.

\cite{wendel:1948} proved: $ \frac{\Gamma(x+y)}{\Gamma(x)} \leq x^y$, for $x>0$ and $y \in [0,1]$. With $x=\frac{D}{2}+k$, $y=\frac{1}{2}$, we obtained $\frac{\Gamma(\frac{D}{2}+\frac{1}{2}+k)}{\Gamma(\frac{D}{2}+k)} \leq \sqrt{\frac{D}{2}+k}$, which leads to: $F(v) \leq \sqrt{2 v^{\top} \Sigma v} \cdot e^{-z} \sum_{k=0}^{\infty} \sqrt{\frac{D}{2}+k} \frac{z^k}{k!}.$

Furthermore, $\sum_{k=0}^{\infty} e^{-z} \frac{z^k}{k!} = 1$ and the function $x \rightarrow \sqrt{\frac{D}{2} + x}$ is concave. Then by Jensen's inequality: $e^{-z} \sum_{k=0}^{\infty} \sqrt{\frac{D}{2}+k} \frac{z^k}{k!} \leq \sqrt{\frac{D}{2}+  e^{-z} \sum_{k=0}^{\infty} \frac{z^k}{k!} k}.$ Knowing that $\sum_{k=0}^{\infty} \frac{z^k}{k!} = z e^{z}$, we have:
$e^{-z} \sum_{k=0}^{\infty} \sqrt{\frac{D}{2}+k} \frac{z^k}{k!} \leq  \sqrt{\frac{D}{2}+ z}.$ \\
And with $z=\frac{\Omega}{2}$, we obtain: $F(v) \leq \sqrt{v^{\top}\Sigma (D+\Omega) v}$.\\

From \cite[p.~442]{muirhead:1984}, the expectation of a non-central Wishart distribution ($G \sim \mathcal{W}_q(D, \Sigma, \Omega)$) is: $\EE[G] = D\Sigma + \Sigma \Omega$. This finally leads to:
\[F(v) \leq \sqrt{v^{\top}\EE[G] v}.\]
\end{proof}

\begin{remark}
As a side note, the second part of the inequality $F(v) \leq \sqrt{v^{\top}\EE[G]v}$ can be obtained using directly Proposition \ref{proposition:sharpenjensen}.
\end{remark}

\coroabsolutebounds*

\begin{proof}
From Proposition \ref{proposition:absolutebounds}, we have $\forall (x,v) \in \mathcal{M} \times {\mathcal T_x M}: \ \sqrt{h(v)} \leq F(v) \leq \sqrt{g(v)}$, with $h:v \rightarrow v^{\top} \alpha  \Sigma v$ and $g:v \rightarrow v^{\top} \EE[G] v$ Riemannian metrics. We also define the parametric curve: $\forall t \in \RR, \gamma(t)=x$ and $\dot{\gamma}(t)=v$. \\

1) Let's show that $L_{\Sigma}(x) \leq L_F(x) \leq  L_R(x)$. Because of the monotonicity of the Lebesgue integrals, we directly have: $\int \sqrt{h(\dot{\gamma}(t))} dt \leq \int F(\dot{\gamma}(t)) dt \leq \int\sqrt{g(\dot{\gamma}(t))} dt$. \\

2) Let's show that $E_{\Sigma}(x) \leq E_F(x) \leq  E_R(x)$. Since all the functions are positive, we can raise them to the power two, and again, with the monotonicity of the Lebesgue integrals, we have: $\int h(\dot{\gamma}(t)) dt \leq \int F^2(\gamma(t),\dot{\gamma}(t)) dt \leq \int\sqrt{g(\dot{\gamma}(t))} dt$. \\

3) Let's show that $V_{\Sigma}(x) \leq V_F(x) \leq  V_R(x)$. 
We write the vectors $v \in {\mathcal T_x M}$ in hyperspherical coordinates: $v=re$, with $r=\norm{v}$ the radial distance and $e$ the angular coordinates. With $v=re$, we have: $r\cdot \sqrt{h(e)} \leq r\cdot F(e) \leq r\cdot g(e) \iff \sqrt{h(e)}^{-1} \geq F(e)^{-1} \geq \sqrt{g(e)}^{-1}$. \\

We want to identify an inequality between the indicatrices, noted $\vol(I_h)$, $\vol(I_g)$, $\vol(I_F)$, formed by the functions $h$, $g$ and $F$. Let's define: $r_g \sqrt{h(e)}=r_h \sqrt{g(e)}=r_F F(e)=1$. For every angular coordinate $e$, we obtain: $r_h \geq r_F \geq r_g$. Intuitively, this means that the finsler indicatrix will always be bounded by the indicatrices formed by $h$ and $g$.  The Busemann-Hausdorf volume of a function $f$ is defined as: $\sigma_B(f) = \vol(\BB^n(1))/\vol(I_f),$ with $\vol(\BB^n(1))$ the volume of the unit ball and $\vol(I_f)$ the volume of the indicatrix formed by $f$.  The previous inequality and the definition of the Busemann-Hausdorff volume implies that: 
$\vol(I_h) \geq \vol(I_F) \geq \vol(I_g) \Rightarrow \sigma_B(h) \leq \sigma_B(F) \leq \sigma_B(g)$. The functions $g$ and $h$ being Riemannian, we have: $\sigma_B(h)=\sqrt{det(\alpha \Sigma)}$ and $\sigma_B(g)=\sqrt{det(\EE[G])}$, which concludes the proof.

\end{proof}

\subsubsection{Relative bounds between the Finsler and the Riemannian metric} \label{appendix:relativebounds}

\propSharpenJensen*
% \begin{lemma} \label{proposition:sharpenjensen}
% Let's note $z=v^{\top} G v$ any random variable. The relative difference between the Finsler metric: $F: (x,v) \rightarrow \EE[\sqrt{v^{\top}G v}]$ and the Riemmanian metric $g: v \rightarrow v^{\top}\EE[G] v$ is:
% \[ 0 \leq \frac{\sqrt{g(v)} - F(v)}{\sqrt{g(v)}} \leq \frac{\Var[z]}{2 \EE[z]^2}. \]
% \end{lemma}

\begin{proof}
We will directly use a sharpen version of Jensen's inequality obtained by \cite{liao:2017}: 
Let X be a one-dimensional random variable with mean $\mu$ and $P(X \in (a,b)) = 1$, where $-\infty \leq a \leq b \leq +\infty$. Let $\phi$ a twice derivable function on $(a,b)$. We further define: $h(x,\mu) = \frac{\phi(x)-\phi(\mu)}{(x-\mu)^2}-\frac{\phi'(\mu)}{x-\mu}$. Then: 
\[\inf_{x \in (a,b)}\{h(x,\mu)\}\Var[X] \leq \EE[\phi(x)] - \phi(\EE[x])\leq \sup_{x \in (a,b)}\{h(x,\mu)\}\Var[X].\]

In our case, we will chose $\phi:z \rightarrow \sqrt{z}$ with $z$ a one-dimensional random variable  defined as $z=v^{\top} G v$. $a=0$, $b=+\infty$ and $\mu=\EE[z]$. $h(z,\mu) = (\sqrt{z}-\sqrt{\mu})(z-\mu)^{-2}-(2(z-\mu)\sqrt{\mu})^{-1}$. Because its first derivative $\phi'$ is convex, the function $x \rightarrow h(x,\mu)$ is monotonically increasing. Thus:

\[\inf_{z \in (0,+\infty)}\{h(x,\mu)\}=\lim_{z\rightarrow 0}=-\frac{\sqrt{\mu}}{2\mu^2} \quad\text{and}\quad \sup_{z \in (0,+\infty)}\{h(x,\mu)\}=\lim_{z\rightarrow +\infty}=0.\]
It finally gives:
\[-\frac{\sqrt{\mu}}{2\mu^2} \Var[z] \leq \EE[\sqrt{z}] - \sqrt{\EE[z]}\leq  0.\]

Replacing $\norm{v}_F \coloneqq F(v)=\EE[\sqrt{z}]$ and $\norm{v}_R \coloneqq \sqrt{g(v)}=\sqrt{\EE[z]}=\sqrt{\mu}$ concludes the proof.
\end{proof}

\begin{lemma} \label{lemma:var(z)/E[z]^2}
Let $z \sim \mathcal{W}_1(D, \sigma, \Omega)$ following a one-dimensional non-central Wishart distribution. Then:
\[ \frac{\Var[z]}{2 \EE[z]^2} = \frac{1}{D+\Omega} + \frac{\Omega}{(D+\Omega)^2}\]
\end{lemma}

\begin{proof}
\cite[Theorem~10.3.7.]{muirhead:1984} states that if $z \sim \mathcal{W}_1(D, \sigma, \omega)$ then
$\EE[z^k] = \sigma^k 2^k \frac{\Gamma(\frac{D}{2} + k)}{\Gamma(\frac{D}{2})}\kummer(-k, \frac{D}{2}, -\frac{1}{2}\Omega)$. In particular, for $k=1$ and $k=2$, we have $\kummer(-1,b,c)=1-\frac{c}{b}$ and $\kummer(-2,b,c)=1-\frac{2c}{b}+\frac{c^2}{b(b+1)}$. We also have $\frac{\Gamma(\frac{D}{2} + 1)}{\Gamma(\frac{D}{2})} = \frac{D}{2}$ and $\frac{\Gamma(\frac{D}{2} + 2)}{\Gamma(\frac{D}{2})} = \frac{D}{2}\left(\frac{D}{2}+1\right)$, which leads to: $\EE[z] = \sigma (D+\Omega)$ and $\EE[z^2] = \sigma^2 (2\omega + 2(D+\omega) + (D+\omega)^2).$ Finally, we conclude:
\[\frac{\Var[z]}{\EE[z]^2}=\frac{\EE[z^2]}{\EE[z]^2}-1=\frac{2\omega}{(D+\omega)^2} + \frac{2}{D+\omega}.\]
\end{proof}

\proprelativebounds*
\begin{proof}
The result is directly obtained using Proposition \ref{proposition:sharpenjensen} and Lemma \ref{lemma:var(z)/E[z]^2}.
\end{proof}

\cororelativebounds*
\begin{proof}
Let's call $M = \max_{v \in {\mathcal T_x M}}\{\frac{\omega}{(D+\omega)^2} + \frac{1}{D+\omega}\}$. 

From Proposition \ref{proposition:relativebounds}, we have: 
\[ 0 \leq \norm{v}_R - \norm{v}_F \leq M\  \norm{v}_R, \quad \text{with} \quad M=\max_{v \in {\mathcal T_x M}}\left\{\frac{1}{D+\omega} + \frac{\omega}{(D+\omega)^2}\right\}\]

1) By the monocity of the Lesbesgue integral, we can directly integrate the previous norms along a curve $\gamma$, which immediately leads to: $0 \leq L_R(x) - L_F(x) \leq M L_R(x)$. \\

2) Since all the functions are positive: $0 \leq \norm{v}_F\leq \norm{v}_R \leq M \norm{v}_R + \norm{v}_F$ leads to: $\norm{v}^2_F\leq \norm{v}_R^2 \leq M^2\norm{v}^2_R + 2M \norm{v}_F\norm{v}_R + \norm{v}^2_F$, and replacing $\norm{v}_F \leq \norm{v}_R$ in the right hand term: $\norm{v}^2_F\leq \norm{v}^2_R \leq (M^2 + 2M) \norm{v}^2_R + \norm{v}^2_F$, and finally: $0 \leq \norm{v}^2_R - \norm{v}^2_F \leq (M^2 + 2M) \norm{v}^2_R$. Again, by continuity of the Lebesgue integral, we directly obtain: $0 \leq E_R(x) - E_F(x) \leq  (M^2 + 2M) E_R(x)$. \\

3) In order to compare the volume between the Finsler and the Riemannian metric, we need to compare the volume of their indicatrices, noted: $\vol(I_g)$ and $\vol(I_F)$ respectively. We write the vectors $v \in {\mathcal T_x M}$ in hypershperical coordinates, with $v=re$, $r=\norm{v}$ the radial distance and $e$ the angular coordinates. The volume of the indicatrices obtained in dimension $q$ (dimension of the latent space) can be written as: $d^{q}V = r^{q-1} dr d\Phi$, with $\Phi$ defining the different angles. We will note $r_F$ and $r_g$ the radial distances of the Finsler and Riemann metrics such that: $\norm{v}_F=r_F \norm{e}_F = 1$ and $\norm{v}_R=r_g \norm{e}_R = 1$ obtained for a specific angle $e$.
 \[\vol(I_F) - \vol(I_g) = \int_{\Phi} \left( \int_{0}^{r_f} r^{q-1} dr - \int_{0}^{r_g} r^{q-1} dr \right) d\Phi = \int_{\Phi} \frac{r_f^{q}}{q} \left(1- \left(\frac{r_g}{r_f}\right)^q \right) d\Phi \leq \int_{\Phi} \frac{r_f^{q}}{q} d\Phi \cdot \left(1- \left(\frac{r_g}{r_f}\right)^q \right),\]
 and by definition: $\vol(I_F) = \int_{\Phi} (r_f^{q}/q) d\Phi$.  Furthermore, for a specific angle $e$, we have: $r_g/r_F = \sqrt{g(e)}/F(e) \geq 1- M$, from Proposition \ref{proposition:relativebounds}. We have: 
 
 \[0 \leq \frac{\vol(I_F) - \vol(I_g)}{\vol(I_F)} \leq  \cdot 1- \left(\frac{r_g}{r_f}\right)^q \leq 1- \left(1- M\right)^q,\]
 and by the definition of the Busemann Hausdorff volume: $ \frac{V_F(x)-V_G(x)}{V_F(x)} = \frac{\vol(I_F)-\vol(I_g)}{\vol(I_F)}$, we conclude the proof.

\end{proof}

\subsubsection{Implications in High Dimensions} \label{appendix:highdimensions}

In this section, we want to show that the difference between the Finsler norm and the Riemannian induced norm, as well as their respective functionals, tend to zero at a rate of $\mathcal{O}(\frac{1}{D})$. We need to be sure that $\omega$ doesn't grow faster than $D$, in other terms: $\omega = \mathcal{O}(D)$. This can be obtained if we assume that every element of the expectation of Jacobian is upper bounded ($\exists m \in \RR^{*}_{+}, \forall i,j \ \EE[J_{ij}] \leq m$). This happens in at least two cases: (1) $\EE[f]$ is somehow Lipschitz continuous; or (2) if $f$ is a Gaussian Process and its covariance is upper bounded. The latter case happens when the process is defined over a bounded domain.\\

\begin{lemma} \label{lemma:omega=O(D)}
Our Finsler metric $v\rightarrow \EE[\sqrt{v^{\top}G v}]$ is defined with $v^{\top} G v \sim \mathcal{W}_1(D, v^{\top} \Sigma v, \omega)$, and $\omega= (v^{\top} \Sigma v)^{-1}(v^{\top} \EE[J]^{\top} \EE[J] v)$. 

If the Finsler manifold is bounded, then: $\omega \leq D M$, with $M\in\RR_{+}$.
\end{lemma}

\begin{proof}
By definition, $\Sigma$ does not depend on $D$. We assume the manifold is bounded, which means that every element of the expected Jacobian is upper bounded: $\EE[J]_{ij} \leq m$, with $m \in \RR^{*}_{+}$. We call $\sigma = v^{\top} \Sigma v \in \RR^{*}_{+}$. 

We have: 
\[\omega = \sigma^{-1} \sum_{k=1}^{D} \sum_{i=1}^{q} \sum_{j=1}^{q} v_i \EE[J]_{ki} \EE[J]_{kj} v_j \leq \sigma^{-1} \sum_{k=1}^{D} m^2 \norm{v}^2 \leq D M,\]
with $M=\sigma^{-1} m^2 \norm{v}^2 \in \RR^{*}_{+}$, and M does not depend on D.
\end{proof}

% {\color{red} Need to finish the proof, the lemma used before was not helpful. We should say that $\omega$ can be potentially lower bounded? Anyway, defined because covariance not equal to zero (otherwise metric equal), and $v^{\top} \EE[J]^{\top} \EE[J] v$ not zero because it would be white noise (and metric would be the same again anyway). 

% Something like: 

% Since we exclude the case where the variance is high, and the case where the expectation of the posterior Jacobian equals zero, we can assume there exist both an upper and lower bound for $v^{\top} \Sigma v$ and $v^{\top} \EE[J]^{\top} \EE[J] v$. In other terms, we can assume that $\omega$ is lower bounded by $m\in\RR^{+}$, and so $\frac{1}{D+\omega} \leq \frac{m}{D}$.

% In addition, we make the hypothesis that the latent space is compact, or at least, we are working on a compact subset of the latent space, ie. we are working on a space that does not contain any holes or missing points. Using the Heine Borel theorem, the metrics and their functionals are bounded. We can now conclude that both metric converge to each other in high dimensions.
% }

\corohighdimfunctionals*

\begin{proof}
From Corollary \ref{corollary:relativebounds}, we directly obtained the results in high dimensions.

We assume that our latent space is bounded, then by \ref{lemma:omega=O(D)}, we have: $0 \leq \omega \leq M D$, with $M\in\RR{+}$.

For the length, we have: 
\begin{equation*}
    \begin{aligned}
        \frac{L_G(x) - L_F(x)}{L_G(x)} & \leq \max_{v \in {\mathcal T_x M}} \left\{\frac{1}{D+\omega} + \frac{\omega}{(D+\omega)^2}\right\} \\
         & \leq  \frac{1+M}{D} 
    \end{aligned}
\end{equation*}

For the energy functional, we have:
\begin{equation*}
    \begin{aligned}
        \frac{E_G(x) - E_F(x)}{E_G(x)} & \leq \max_{v \in {\mathcal T_x M}} \left\{\frac{2}{D+\omega} + \frac{1+2\omega}{(D+\omega)^2} + \frac{2\omega}{(D+\omega)^3} + \frac{\omega^2}{(D+\omega)^4} \right\} \\
         & \leq \frac{2+2M}{D} + \frac{1+2M+M^2}{D^2} \\
        \limsup\limits_{D\rightarrow\infty}{D \times \frac{E_G(x) - E_F(x)}{E_G(x)}} & \leq \limsup\limits_{D\rightarrow\infty} {2(1+M)+\frac{M^2+2M+1}{D^2}} \to 2(1+M)\\
    \end{aligned}
\end{equation*}

For the volume, we have:
\begin{equation*}
    \begin{aligned}
        \frac{V_G(x) - V_F(x)}{V_G(x)} & \leq 1 - \left(1-\max_{v\in{\mathcal T_x M}} \left\{\frac{1}{D+\omega} + \frac{\omega}{(D+\omega)^2} \right\}\right)^{q} \\
         & \leq 1 - \left( 1 - \frac{1+M}{D}\right)^{q} 
    \end{aligned}
\end{equation*}

Using Taylor series expansion, when $x \sim 0$, we have: $1 - \left( 1 - x\right)^{q} = qx + o(x^2)$. Let's call $\varepsilon \ll 1$, and rewrite the Taylor series:  

\begin{equation*}
    \begin{aligned}
        \frac{V_G(x) - V_F(x)}{V_G(x)} & \leq q \frac{1+M}{D} + \varepsilon q \frac{1+M}{D} \\
        \limsup\limits_{D\rightarrow\infty}{\frac{D}{q}\times \frac{V_G(x) - V_F(x)}{V_G(x)}} & \leq (1+M)(1+\varepsilon)
    \end{aligned}
\end{equation*}

The difference between the functionals can converge to zero if they are similar in high dimensions, or if they all diverge to infinity. This latter case does not happen as we assume the latent manifold being bounded, and so the metrics are then finite, which concludes the proof.

\end{proof}

\corohighdimmetrics*
\begin{proof}
Similar to the \ref{corollary:highdimensionsfunctionals}, assuming that our latent space is bounded, from \ref{lemma:omega=O(D)}, we have $0 \leq \omega \leq M D$. From \ref{proposition:relativebounds}, we deduce:

\begin{equation*}
    \begin{aligned}
        0 \leq \frac{\norm{v}_R - \norm{v}_F}{\norm{v}_R} & \leq \frac{1}{D+\omega} +\frac{\omega}{(D+\omega)^2}\\
         & \leq  \frac{1+M}{D} 
    \end{aligned}
\end{equation*}

In a bounded manifold, the metric are finite. We can deduce that they converge to each other in high dimensions.
\end{proof}
\section{Experiments}
\label{appendix:experiments}
\subsection{Datasets}

\subsubsection{Font data}
The dataset represents 46 different font for each letter (upper and lower case) whose contour is parametrised by a spline (or two splines, depending on the letter used) obtained from at least 500 points \cite{campbell:2014}. 

In our case, we choose to learn the manifold of the letter $\textbf{f}$. The dataset is composed of 46 different fonts, each letter being drawn by 1024 points. We reduce this number from 1024 to 256 by sampling one point every 4. The dimension of the observational space is then 256. 

\subsubsection{qPCR}

The qPCR data, gathered from \cite{Guo:2010}, was used to illustrate the training of a GPLVM in Pyro \cite{pyro:qPCR} and is available at the Open Data Science repository \cite{ahmed:2019}. It consists of 437 single-cell qPCR data for which the expression of 48 genes has been measured during 10 different cell stages. We then have 437 data points, 48 observations, and 10 classes. 
Before training the GP-LVM, the data is grouped by the capture time, as illustrated in the Pyro documentation.  

\subsubsection{Pinwheel on a sphere}

A pinwheel in 2-dimension is created and then projected onto a sphere using a stereographic projection method. The final dataset is composed of 1000 points with their coordinates in 3-dimensions.

\subsection{GP-LVM training}
We learn our two-dimensional latent space by training a GP-LVM \cite{lawrence:2003} with Pyro \cite{bingham:pyro:2019}. The Gaussian Process used is a Sparse GP, defined with a kernel (RBF, or Matern) composed of a one-dimensional lengthscale and variance. The parameters are learnt with the Adam optimiser \cite{kingma:2014}. The number of steps and the initialisation of the latent space vary with the dataset.

\begin{table}[htbp]
\centering
\begin{tabular}{lllll}
\hline
datasets                 & pinwheel   & font data & qPCR         \\ \hline
Number of data points    & 500       & 46        & 437          \\
Number of observations   & 3          & 256       & 48           \\ \hline
initialisation           & PCA        & PCA       & custom       \\ 
kernel                   & RBF        & Matern52  & Matern52     \\
steps                    & 17000      & 5000      & 5000         \\ 
learning rate            & 1e-3       & 1e-4      & 1e-4         \\ \hline
lengthscale              & 0.24       & 0.88      & 0.15         \\
variance                 & 0.95       & 0.30      & 0.75         \\
noise                    & 1e-4       & 1e-3      & 1e-3         \\ \hline

\end{tabular}
\caption{Description of the datasets trained with a GP-LVM.}
\end{table}

\subsection{Computing indicatrices}
\label{appendix:experiments:subsec:indicatrices}
An indicatrix of a function $g$ at a point $x$ is defined such that: $v \in {\mathcal T_x M} | g_x(v) < 1$. In other terms, the indicatrix is the representation of a unit ball in our latent space. 
If we use an euclidean metric, our indicatrix in our 2-dimensional latent space would be a unit ball, as we need to solve: $v \in {\mathcal T_x M}$, $\norm{v} < 1$. For a Riemannian metric, our indicatrix is necessarily an ellipse, whose semi axis are the square-roots of the eigenvalues of the metric tensor $G$:: $v \in {\mathcal T_x M}$, $v^{\top} G v < 1$. For our Finsler metric, we don't have an analytical solution, and so it's difficult to predict the shape of the convex polygon.

In this paper, the indicatrices are drawn the following way: for a single point in our latent space, we compute the value of $v^{\top} G v$ and $F(x,v)$ for $v$ varying over the space. We then extract the contour when  $v^{\top} G v$ and $F(x,v)$ are equal to 1. Computing the area of the indicatrices will be used in the  section \ref{appendix:experiments:volume} to compute the volume measures.

\subsection{Computing the volume forms}
\label{appendix:experiments:volume}
For the figures used in this paper, by default, the background of the latent space represents the volume measure of the expected Riemannian metric ($V_{G} = \sqrt{\EE [G]}$) on a logarithm scale. In figure \ref{fig:volume_measures}, the volume measure of the Finsler metric is also computed. 

\subsubsection{Finsler metric}
To compute the volume measure of our Finsler metric, we choose the Busemann-Hausdorff definition, which is the ratio of a unit ball over the volume of its indicatrix: $\mathcal{V} = \vol(\BB^n(1))/\vol(\{v \in {\mathcal T_x M} | F(x,v) < 1\})$. While our Finsler function has an analytical form, its expression doesn't allow to directly solve the equation: $v \in {\mathcal T_x M}, F(x,v) < 1$. Instead we approximate its indicatrix as describe in \ref{appendix:experiments:subsec:indicatrices}, using a contour plot and extracting the paths vertices. We can then compute the area of the obtained polygon, and divide with the volume of a unit ball: $\vol(\BB^2(1)) = \pi$. 

The volume measure can then be computed for each point over a grid (32 x 32, in figure \ref{fig:volume_measures}), and we interpolate all the other points. Note that this method can only be used when our latent space is of dimension 2.

\subsubsection{Expected Riemannian metric}
There is two ways to compute the volume measure of the expected Riemannian metric. One way is to directly use the metric tensor: $V_{G} = \sqrt{\EE [G]}$. Another one is to remember that any Riemannian metric is a Finsler metric, and thus, the Busemann-Hausdorff definition also applied for our metric: $V_{G} = \vol(\BB^n(1))/\vol(\{v \in {\mathcal T_x M} | v^{\top} \EE [G] v < 1\})$. Solving $v^{\top} \EE [G] v < 1$ for $v \in {\mathcal T_x M}$ is equivalent to solving the area of an ellipse.

For the first method, we can either sample multiple times the metric, which is computationally expensive, or use the fact that our metric tensor is a non-central Wishart matrix: $G = J^{\top}J \sim \mathcal{W}_q(D, \Sigma, \Sigma^{-1} \EE[J]^{\top} \EE[J])$, with $\Sigma$ the covariance of the Jacobian $J$ and $D$ the dimension of the observational space. In this case, its expectation is: $\EE[G] = \EE[J]^{\top} \EE[J] + D \Sigma$. We can access the derivatives of the function $f$ (detailed in section \ref{appendix:computation:derivatives}), and compute both quantities $\EE[J]$ and $\Sigma$ needed to estimate the expected metric and its determinant.

For the second method, we can compute the area of the ellipse in the same way we compute the Finsler volume measure.

\subsection{Experiments when increasing the number of dimensions}
In Figure \ref{fig:highdims}, we computed the volume ratio and draw indicatrices while varying the number of dimensions, to illustrate that both our Finsler metric and the expected Riemannian metric seem to converge when $D$ increases. 

The main issue with this experiment is to vary only one factor, the number of dimensions $D$, while keeping the other factors unchanged. This is difficult for two reasons: 1) Even with a very low dimensional observational, both metrics are already very similar to each other. It would be difficult to illustrate a convergence while increasing the number of dimensions. 2) the function $f$ needs to be learnt again each time we increase the number of dimensions of the observational space, and the parameters of the Gaussian Process will change too. 

Instead, we try to illustrate our results by computing empirically the stochastic metric tensor $G = J^{\top}J$, using its Jacobian $J \sim \mathcal{N}(\EE[J],\Sigma)$, a $D \times q$ matrix. The number of dimensions is modified by simply truncatenating the Jacobian $J$. In Figure \ref{fig:highdims}, the volume ratio is computed for 12 Jacobians obtained with different random parameters $\EE[J]$ and $\Sigma$. The Finsler and Riemannian indicatrices (lower right) are drawn for only one Jacobian selected randomly.
\section{Computations}
\subsection{Computing geodesics with Stochman and minimising the curve energy functionals}
 
 An essential task is to compute shortest paths, or geodesics, between data points in the latent space. Those shortest paths can be obtained in two ways: either by solving a corresponding system of ODEs, or by minimising the curve energy in the latent space. The former being computationally expensive, we favour the second approach which consists in optimising a parameterised spline on the manifold. This method is already implemented in Stochman \cite{software:stochman}, where we can easily optimise splines by redefining the curve energy function of a manifold class.
 
 We need two curvge energy functional: one for the expected Riemannian metric and one for the Finsler metric.
 
 %By definition, the minimum of the length should be used as the objective function, but this leads to an unstable optimisation. The length functional being reparametrisation invariant, the solver can indeed find an infinite number of solutions. A clever hack is to minimise the energy, which is a convex function and an upper bound of the length: $\mathrm{Length}^2(\gamma) \leq 2\mathrm{Energy}(\gamma)$.
 \subsubsection{Curve energy for the Riemannian metric}
 We know that the stochastic metric tensor $G_t$ defined on a point $t$ follows a non-central Whishart distribution. Thus, we can compute its expectation $\EE[G_t]$ knowing the Jacobian covariance and expectation: $\EE[J_t]$ and $\Sigma$. The next Section \ref{appendix:computation:derivatives} explains how to compute those quantities. 
 
 Assuming the spline is discretized into N points, we can compute the curve energy with:
 \[
E_{G}(\gamma(t)) = \int_0^1 \dot{\gamma}(t)^{\top} \EE[G_t]\dot{\gamma}(t) \,dt \approx \sum_{i=1}^{N} \dot{\gamma}_i^{\top} \left(\EE[J_i]^{T} \EE[J_i] + D \Sigma_i \right) \dot{\gamma}_i.
\]
 
\subsubsection{Curve energy for the Finsler metric}
In order to compute of the curve energy $\mathcal{E}(\gamma)$, we must first derive the expectation $\EE[J_t]$ and covariance $\Sigma$ of the Jacobian of $f$, which should follow a normal distribution: $J_i = \partial f_{i} \sim \mathcal{N} \left(\EE[J], \Sigma \right)$. We assume the points $\partial f_{i}$ are independent samples with the same variance drawn from a normal distribution. We can then compute the Finsler metric which follows a non-central Nakagami distribution (See Proposition \ref{proposition:definition:finsler}):

\[
\mathcal{E}(\gamma(t)) = \int_0^1 F(t, \dot{\gamma}(t))^2 \,dt \approx \sum_{i=1}^{N} 2 \dot{\gamma}_i^{\top} \Sigma_i \dot{\gamma}_i \left( \frac{\Gamma(\frac{D}{2} + \frac{1}{2})}{\Gamma(\frac{D}{2})} \right)^2 \kummer\left(-\frac{1}{2}, \frac{D}{2}, -\frac{\omega_i}{2}\right)^2,
\]
with $\kummer$ the confluent hypergeometric function of the first kind and $\omega_i= (\dot{\gamma}_i^{\top} \Sigma_i \dot{\gamma}_i)^{-1}(\dot{\gamma}_i^{\top} \Omega_i \dot{\gamma}_i)$ and $\Omega_i = \Sigma_i^{-1} \EE[J_i]^{\top}\EE[J_i]$.

This function has been implemented in Pytorch using the known gradients for the hypergeometric function: $\frac{\partial}{\partial x} \kummer(a,b,x) = \frac{a}{b} \kummer(a+1, b+1,x)$.

\subsection{Accessing the posterior derivatives}
\label{appendix:computation:derivatives}

We assume that the probabilitic mapping $f$ from the latent variables $X$ to the observational variables $Y$ follows a normal distribution. We would like to obtain the posterior kernel $\Sigma_{*}$ and expectation $\mu_{*}$ such that $p(\partial_{t} f|Y,X) \sim \mathcal{N} \left(\mu_{*}, \Sigma_{*}\right)$. 

We make the hypothesis the observed variables are modelled with a gaussian noise $\epsilon$ whose variance is the same in every dimension. In particular, for the $n^{th}$ latent ($x$) and observed ($y$) variable in the $j^{th}$ dimension: $y_{n,j} = f_{j}(x_{n,:})+\epsilon_{n}$. Thus, the output variables have the same variance, and the posterior kernel $\Sigma_{*}$ is then isotropic with respect to the output dimensions: $\Sigma_{*} = \sigma^2_{*} \cdot I_{D}$. 

There are two ways of obtaining the posterior variance and expectation:
\begin{itemize}
  \item  We use the gaussian processes to predict the derivative ($\partial_{c} f $) of the mapping function $f$, and we multiply the obtained posterior kernel by the curve derivative ($\partial_{t}c$), following the chain rule: $\frac{df(c(t))}{dt} = \frac{df}{dc} \cdot  \frac{d c}{dt}$ (Section: \ref{subsec: closed-form})
  \item We discretize the derivative of the mapping function as the difference of this function evaluated at two close points. We use a linear operation to obtain the posterior variance and expectation: $\partial_{t}f(c(t)) \sim f(c(t_{i+1})) - f(c(t_{i})) $. (Section: \ref{subsec:discretization})
\end{itemize}

\subsubsection{Closed-form expressions}
\label{subsec: closed-form}

We assume that $f$ is a Gaussian process. Hence, because the differentiation is a linear operation, the derivative of a Gaussian Process is also a Gaussian Process \cite{rasmussen:gaussian:2006}. 

The data $Y \in \mathcal{R}^{N \times D}$ follows a normal distribution, so we can infer the partial derivative of one data point $(J^{T})_{ji} = \frac{\partial y_i}{\partial x_j}$, with $i = 1 \ldots D$ and $j = 1 \ldots d$. We have:

\[
  \left[\begin{array}{c}  Y \\ (J)^{\top} \end{array}\right]
  =
  \prod_{i=1}^{D} \mathcal{N}
  \left(
    \left[\begin{array}{c} \mu_{y} \\ \mu_{\partial y} \end{array}\right],
    \left[\begin{array}{cc} K(x, x) & \partial K(x, x_{*}) \\
                            \partial K(x_{*}, x) & \partial^2 K(x_{*}, x_{*})
          \end{array} \right]
  \right).
\]

and $J^{\top}$ can be predicted:
\[
  p(J^{\top} | Y, X) = \prod_{i=1}^{D} \mathcal{N} \left( \mu_{*}, \Sigma_{*} \right),
\]

with:
\begin{align*}
  \mu_{*} = \partial K(x_{*}, x) \cdot K(x, x)^{-1} \cdot (y - \mu_y) + \mu_{\partial y}\\
  \Sigma_{*} = \partial^2 K(x_{*}, x_{*}) - \partial K(x_{*}, x) \cdot K(x, x)^{-1} \cdot \partial K(x, x_{*}).
\end{align*}

Finally, $\partial_t f$ is obtained:
\[
  p(\partial_t f(c(t)) | f(x), x) = \prod_{i=1}^{D} \mathcal{N} \left(\dot{c} \mu_{*}, \dot{c}^{\top}\Sigma_{*}\dot{c} \cdot I_{D}\right).
\]

\subsubsection{Discretization}
\label{subsec:discretization}
One can notice that: $\partial_t f(c(t)) \sim f(c(t_{i+1})) - f(c(t_{i}))$. We know that $f(c(t_{i+1}))$ and $f(c(t_{i}))$ both follows a normal distribution.

\[
  \left[\begin{array}{c}  f(c(t_{i})) \\ f(c(t_{i+1})) \end{array}\right]
  = \prod_{j=1}^{D} \mathcal{N}
  \left(
    \left[\begin{array}{c} \mu_{i} \\ \mu_{i+1} \end{array}\right],
    \left[\begin{array}{cc} \sigma^2_{ii} & \sigma^2_{i,i+1} \\
                            \sigma^2_{i+1, i} & \sigma^2_{i+1,i+1}
          \end{array} \right]
  \right).
\]

If $Y = AX$ affine transformation of a multivariate Gaussian $X \sim \mathcal{N}(\mu, \sigma^2)$, then Y is also a multivariate Gaussian with: $Y \sim \mathcal{N}(A \mu, A^T \sigma^2 A)$. In our case, we choose $A^T = [-1, 1]$. We have:

\[
f(c(t_{i+1})) - f(c(t_{i})) \sim  \mathcal{N}(\mu_{*}, \sigma_{*}^2 \cdot I_{D}),
\]

with:
\begin{align*}
  \mu_{*} =  \mu_{i+1} - \mu{i} \\
  \sigma_{*}^2  = \sigma^2_{ii} + \sigma^2_{i+1, i+1} - 2 \sigma^2_{i,i+1}
\end{align*}

\begin{figure}[htbp]
    \centering
    \includegraphics[width=0.7\textwidth]{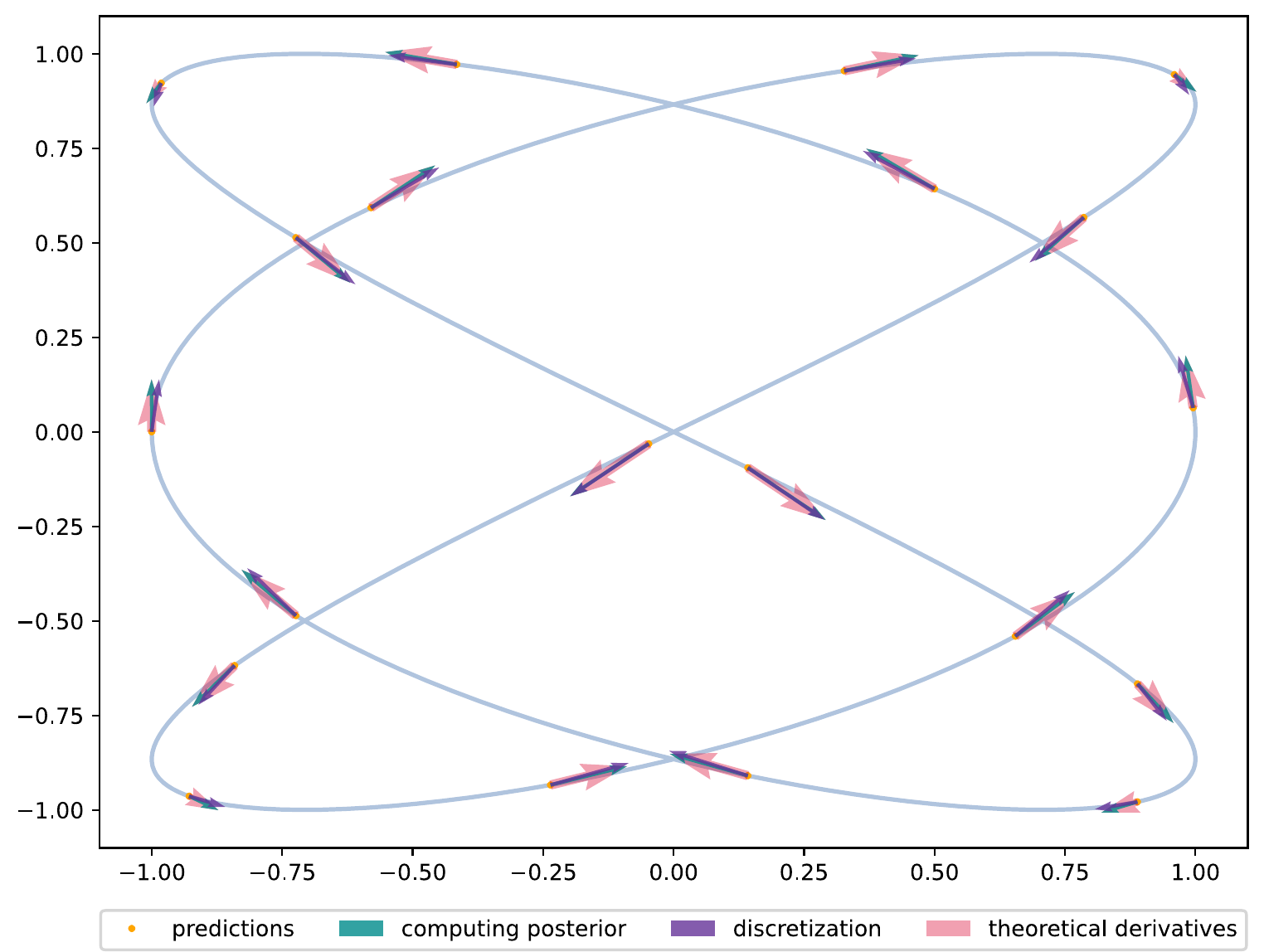}
    \caption{Illustration of the derivatives obtained with a trained GP on a simple parametrised function: both methods give the correct derivatives if enough points are sampled.}
    \label{fig:appendix:derivatives}
\end{figure}

\end{document}